\documentclass[lettersize,journal]{IEEEtran}
\usepackage{amsmath,amsfonts}
\usepackage{algorithm}
\usepackage{array}
\usepackage[caption=false,font=normalsize,labelfont=sf,textfont=sf]{subfig}
\usepackage{textcomp}
\usepackage{stfloats}
\usepackage{url}
\usepackage{verbatim}
\usepackage{graphicx}
\usepackage{cite}
\usepackage{booktabs}
\usepackage{algpseudocode}
\usepackage{afterpage}
\usepackage{lipsum}
\usepackage{float}
\usepackage{multicol} 
\usepackage{xspace}

\usepackage{algorithm}
\usepackage{amsmath}
\usepackage{amssymb}
\usepackage{multirow}
\usepackage{booktabs}
\usepackage{makecell}
\usepackage[table,xcdraw]{xcolor}
\begin{document}

\title{Denoise-I2W: Mapping Images to Denoising Words for Accurate Zero-Shot Composed Image Retrieval}

\author{Yuanmin Tang, Jing Yu*, Keke Gai, Jiamin Zhuang, Gaopeng Gou, Gang Xiong, Qi Wu
\thanks{This work was supported by the Central Guidance for Local Special Project (Grant No.Z231100005923044) and the Climbing Plan Project (Grant No E3Z0261)}
\thanks{Yuanmin Tang, Jiamin Zhuang, Gaopeng Gou, and Gang Xiong are with the Institute of Information Engineering, Chinese Academy of Sciences, China, and the School of Cyber Security, University of Chinese Academy of Sciences, China. (e-mail: tangyuanmin@iie.ac.cn; yujing02@iie.ac.cn; gougaopeng@iie.ac.cn, xionggang@iie.ac.cn)}
\thanks{Jing Yu is with the School of Information Engineering, Minzu University of China, Beijing, China
jing.emy.yu01@gmail.com)}
\thanks{Keke Gai is with the School of Cyberspace Science and Technology, Beijing Institute of Technology, Beijing, China. (e-mail: gaikeke@bit.edu.cn)}

\thanks{Qi Wu is with the Australia Centre for Robotic Vision (ACRV), the University of Adelaide. (e-mail:qi.wu01@adelaide.edu.au)}
\thanks{Corresponding author: Jing Yu (e-mail: jing.emy.yu01@gmail.com).}
\thanks{This work has been submitted to the IEEE for possible publication. Copyright may be transferred without notice, after which this version may no longer be accessible.}
}

\markboth{Journal of \LaTeX\ Class Files,~Vol.~14, No.~8, August~2021}%
{Shell \MakeLowercase{\textit{et al.}}: A Sample Article Using IEEEtran.cls for IEEE Journals}


\maketitle

\begin{abstract}

Zero-Shot Composed Image Retrieval (ZS-CIR) supports diverse tasks with a broad range of visual content manipulation intentions that can be related to domain, scene, object, and attribute. A key challenge for ZS-CIR is to accurately map image representation to a pseudo-word token that captures the manipulation intention relevant image information for generalized CIR. However, existing methods between the retrieval and pre-training stages lead to significant redundancy in the pseudo-word tokens. In this paper, we propose a novel denoising image-to-word mapping approach, named Denoise-I2W, for mapping images into denoising pseudo-word tokens that, without intention-irrelevant visual information, enhance accurate ZS-CIR. Specifically, a pseudo triplet construction module first automatically constructs pseudo triples (\textit{i.e.,} a pseudo-reference image, a pseudo-manipulation text, and a target image) for pre-training the denoising mapping network. Then, a pseudo-composed mapping module maps the pseudo-reference image to a pseudo-word token and combines it with the pseudo-manipulation text with manipulation intention. This combination aligns with the target image, facilitating denoising intention-irrelevant visual information for mapping. Our proposed Denoise-I2W is a model-agnostic and annotation-free approach. It demonstrates strong generalization capabilities across three state-of-the-art ZS-CIR models on four benchmark datasets. By integrating Denoise-I2W with existing best models,  we obtain consistent and significant performance boosts ranging from 1.45\% to 4.17\% over the best methods without increasing inference costs. and achieve new state-of-the-art results on ZS-CIR. Our code is available at \url{https://github.com/Pter61/denoise-i2w-tmm}.

\end{abstract}


\section{Introduction}

Distinct from traditional image retrieval \cite{datta2008image}, Composed Image Retrieval (CIR) \cite{vo2019composing} aims to retrieve images similar to a reference image with modification according to a manipulated text description. Considering both visual and textual query intent, CIR becomes more applicable in internet search and e-commerce due to its flexibility. Several CIR tasks have been proposed, including scene image search by object composition and manipulation, fashion image search by attribute changes, and creative search by image style conversion\cite{Chen_2020_CVPR,Saito_2023_CVPR}. The key challenges of CIR lie in two folds: (1) Understanding the manipulation intention in the text description and accurately representing the corresponding visual content in the reference image, and (2) bridging the heterogeneous gap between the manipulation intention in textual modality and the reference information in visual modality, and combining them without irrelevant noise for accurate target image retrieval.

\begin{figure}[t]
    \centering
    \includegraphics[width=1.0\linewidth]{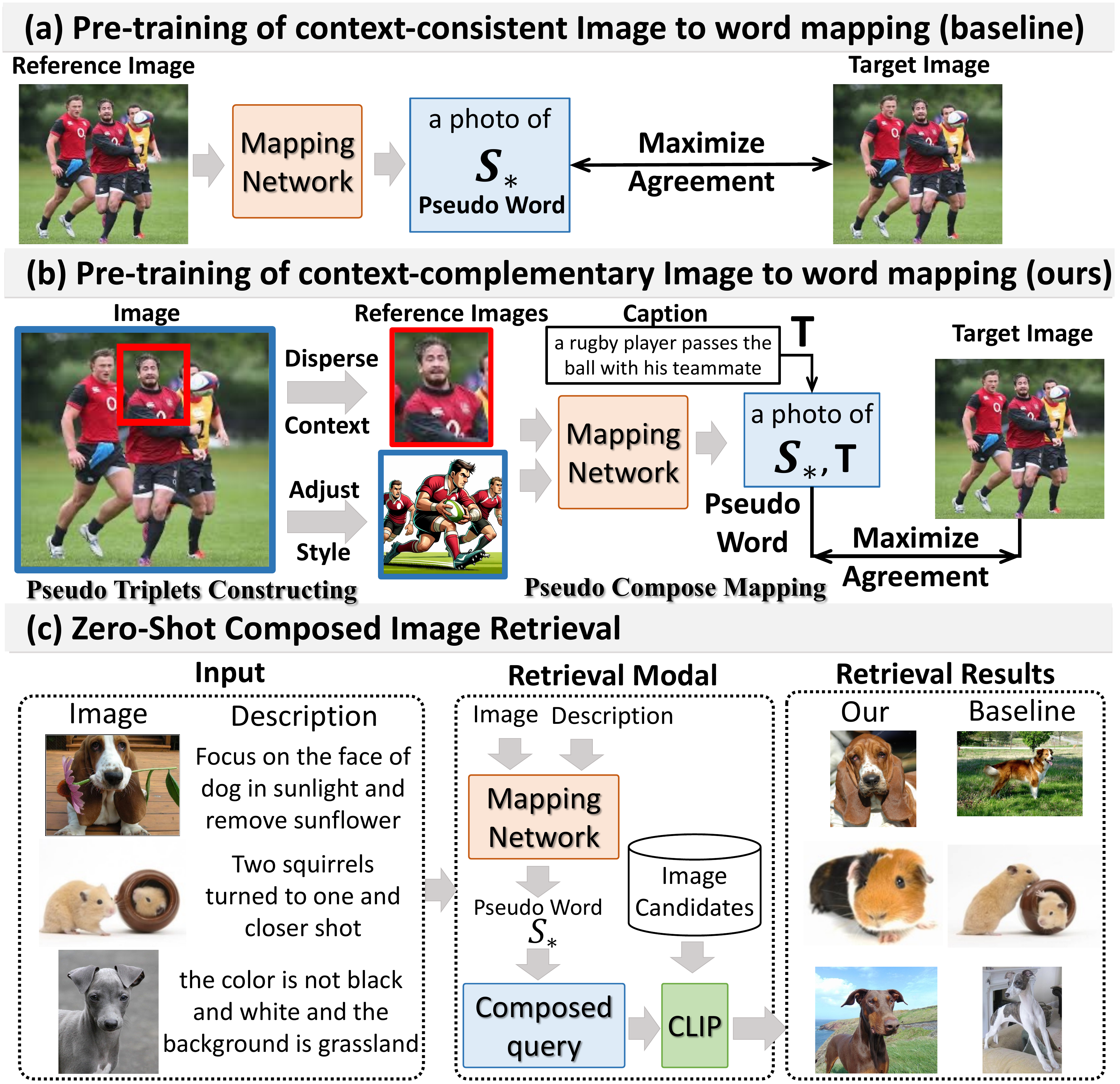}
    \caption{An illustration of our motivation. (a) Noising context-consistent visual mapping. (b) Our denoising context-complementary visual mapping. (c) ZS-CIR process with the results of different mapping strategies.} 
    \label{fig:motivation}
\end{figure}

To solve the above challenges, various supervised methods have been proposed \cite{Chen_2020_CVPR,Liu_2021_ICCV,Goenka_2022_CVPR,Baldrati_2022_CVPR}, which requires a great amount of annotated triplets, \textit{i.e.,} a reference image, a manipulation text description, and a target image, for training task-specific retrieval models. These supervised methods are labor-consuming for data annotation and hard for model generalization. Recent progress \cite{Saito_2023_CVPR,baldrati2023zero,tang2023contexti2w} has led to the study of Zero-Shot Composed Image Retrieval (ZS-CIR), which aims to conduct CIR tasks without relying on annotated triplets for training. Existing solutions for ZS-CIR map a reference image to the language space, combining it with a manipulation text to form a query. he query retrieves target images by calculating their semantic similarity in a common semantic space of a pre-trained vision-language model by calculating semantic similarity. These methods typically involve a pre-trained mapping network that converts the reference image into a pseudo-word token, denoted as $\boldsymbol{S}_*$. During retrieval, $\boldsymbol{S}_*$  is combined with the manipulation text to construct a query, which a pre-trained CLIP model \cite{radford2021learning} then encodes. Then, these methods leverage the pre-trained vision-language models to form a common semantic space and measure the similarity between the formed query and the target images in that space. Despite the significant progress, the retrieval performance is not satisfactory because there exist gaps between the pre-training stage and the retrieval stage:

(1) The desired visual information differs between image-to-word mapping and composed image retrieval. Existing pre-training methods map the whole image information to a pseudo-word token $S_{*}$ regardless of possible human manipulation intention, as shown in Figure \ref{fig:motivation}(a). Without specific mapping intent, $S_{*}$ contains heavy information redundancy involving most objects, background/foreground, color, style, and texture. During retrieval, human has some typical manipulation intention (\textit{e.g.,} changes on objects, attributes, and styles) that focuses on specific visual aspects and local content, which is difficult to capture from the information-dense and highly entangled pseudo-word token $S_{*}$. Considering the queries in Figure \ref{fig:motivation}(c), existing pre-trained $S_{*}$ fails when the description manipulates on fine-grained visual content, such as visually overlapped objects (\textit{e.g.},
dog behind the sunflower), multiple objects
(\textit{e.g.}, one of the two squirrels) and the specific image style (\textit{e.g.}, the image color).

(2) A discrepancy exists between the retrieval and pre-training stages in ZS-CIR models. During retrieval, the mapping network is tasked with aligning intention-relevant visual information (\textit{e.g.,} domain, scenes, objects, and attribute) in language space to form a composed image description query (\textit{e.g.,} change to a woman playing the violin joyfully in the street). Then, the model calculates semantic similarity with the target image. However, in the pre-training phase, the mapping network is trained to align general visual information with textual descriptions of the image content (\textit{e.g.,} a musician plays the piano). Without training with manipulation text, the network makes it hard to select the intent-specific visual information for mapping, leading to inaccurate retrieval.

To tackle these issues, we consider the human manipulation intention in the pre-training phase and propose an intention-based \textbf{Denois}ing mapping approach to convert \textbf{I}mages into pseudo \textbf{W}ords (\textbf{Denoise-I2W}), which reduces redundant visual information and focuses on the visual aspects of human manipulation intention for accurate ZS-CIR. We design a pseudo composed image retrieval pre-training training scheme as illustrated in Figure \ref{fig:motivation}(b): the Pseudo Triplet Construction module first automatically constructs pseudo $<$reference image, manipulation text, target image$>$ triplets based on the off-the-shelf image-caption pairs. To be generic for diverse downstream CIR tasks, we propose two reference image selection strategies of image cropping and style transfer, which imitate both object-level (\textit{e.g.} foreground/background differentiation, multiple object selection, and object attribute editing) and style-level (\textit{e.g.} color changes and domain conversion) CIR tasks.  Then, the Pseudo Composed Mapping module converts the pseudo reference image into a pseudo-word token and combines it with the target image caption to align with the target image. Without the need for expensive triplet annotations, Denoise-I2W is trained on pseudo-triplets that are automatically constructed on public image-caption data and are easily integrated with various ZS-CIR models.

The main contributions are summarized as follows: 

\begin{itemize} \item[$\bullet$] We introduce a novel method for automatically constructing pseudo-triplets using existing image-caption data. These pseudo-triplets are effectively applied to diverse CIR tasks, enabling efficient pre-training for vision-to-language alignment. \item[$\bullet$] We propose a denoising-based pre-training approach that leverages pseudo-triplets to remove irrelevant visual information, focusing on the manipulation intent. This refinement enhances retrieval accuracy by aligning visual and textual modalities more accurately. \item[$\bullet$] Our method is model-agnostic, meaning the pre-trained alignment module can seamlessly integrate with existing ZS-CIR models. Without modifying the model architecture or requiring additional annotated data, Denoise-I2W consistently improves retrieval performance across multiple CIR tasks, demonstrating its efficiency and effectiveness. \end{itemize}

\section{Related Works}

\begin{figure*}
    \centering
    \includegraphics[width=1.00\linewidth]{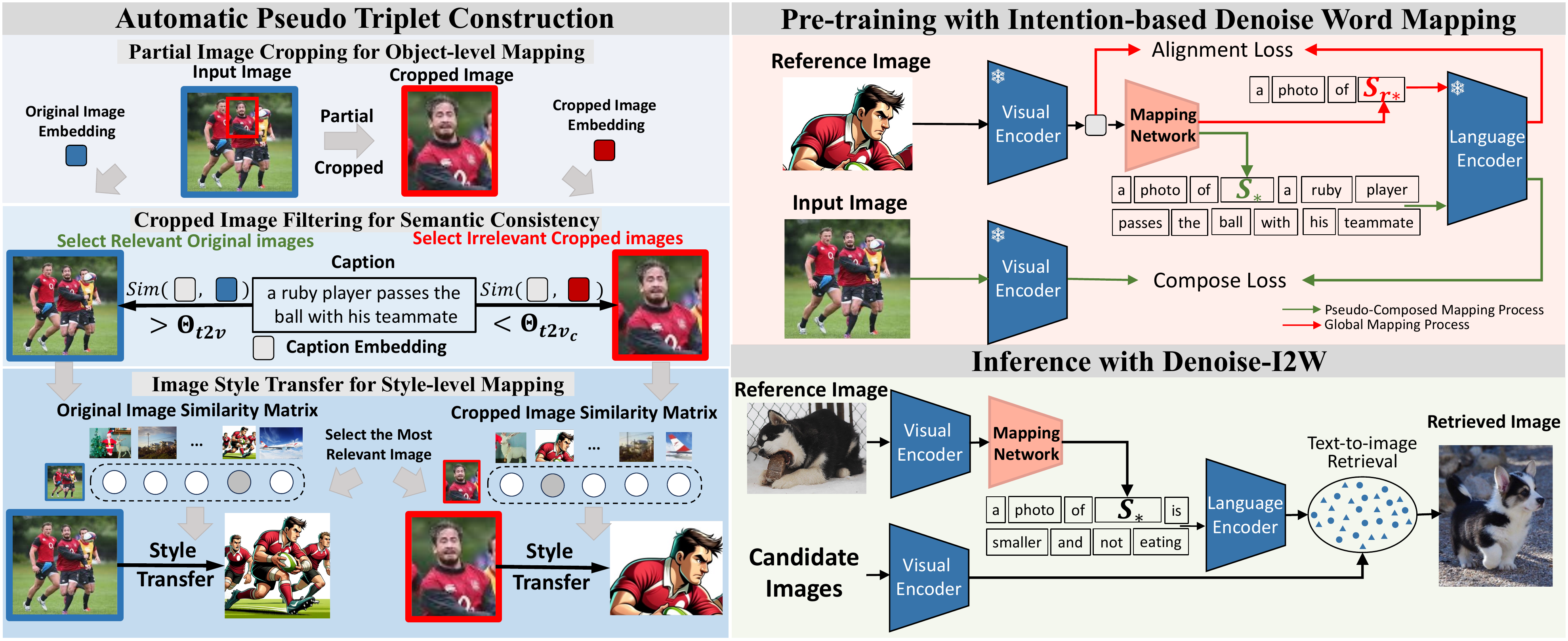}   
    \caption{An overview of our Denoise-I2W. Pseudo Triplet Construction (left): Automatically constructing pseudo triples with partial-cropped images and style-transfer images from existing image-caption pairs. Pre-training and Inference (right): Mapping the intention-based denoising visual content to a pseudo-word token $S_*$ and forming the composed query in a unified language space for ZS-CIR.}
    \label{fig:model-architecture}
\end{figure*}

\noindent \textbf{Composed Image Retrieval} Composed Image Retrieval (CIR) integrates image and textual data for retrieval purposes \cite{Vo_2019_CVPR}. Existing models primarily employ late-fusion techniques to merge separate visual and linguistic features \cite{Baldrati_2022_CVPR, Liu_2021_ICCV}. In contrast, zero-shot CIR methods \cite{Saito_2023_CVPR, baldrati2023zero, gu2024lincir, zhang2024magiclens, agnolucci2024isearle, xu2024set} such as Pic2Word \cite{Saito_2023_CVPR}, SEARLE \cite{baldrati2023zero}, and Context-I2W \cite{tang2023contexti2w} train on standard image-text pairs to overcome the need for extensive CIR datasets. Specifically, Pic2Word correlates complete images with textual features, SEARLE incorporates a pseudo-word token into GPT-based captions, and Context-I2W utilizes context-dependent word mapping for enhanced retrieval accuracy. However, these models generally encode entire images into a single pseudo-word token without accounting for manipulation intention during pre-training, introducing significant noise and inaccuracies in query formulation. To mitigate this, we propose an intention-based denoising word mapping approach that allows the text encoder to selectively engage with intention-relevant image features, thus refining retrieval accuracy. 

Several methods aim to lessen the task mismatch in zero-shot CIR. Chen and Lai \cite{mtcir} propose a triplet synthesis comprising an original image, its corresponding caption, and a masked image, treating the original as the target, the caption as manipulation text, and the masked image as a reference. Despite this innovative approach, discrepancies persist between the manipulation text (\textit{e.g.,} ``change the dog to a cat'') and image captions (\textit{e.g.,} ``a dog jumps to catch a frisbee''), necessitating extensive fine-tuning of the CLIP model, which alters the visual embeddings within the retrieval database. In contrast, our method leverages existing image-caption datasets for training without disrupting the CLIP model's established knowledge base. Moreover, CIReVL \cite{training-free-cir} attempts to bridge task discrepancies by generating descriptive captions for composed queries using a sophisticated captioning model and a large language model (LLM). While CIReVL demonstrates impressive performance without training, it requires costly and inefficient image captioning inferences \cite{blip,blip2}, and successful implementation depends on finely tuned, task-specific prompts. Our approach surpasses CIReVL in efficiency and operates autonomously without requiring manual intervention.

\noindent \textbf{Vision and Language Pre-training Models.} Typical Vision and Language Pre-training (VLP) models like CLIP \cite{radford2021learning} leverage large-scale image-text pair training, achieving implicit image-text alignment. Recent VLP advancements \cite{Zhou_2022_CVPR,song2022clip,clip2fl,mmm} utilize frozen models to map encoded image and text features into a unified semantic space, facilitating various zero-shot tasks \cite{pmlr-v162-li22n,song2022clip}. Nonetheless, current zero-shot learning approaches based on CLIP, which is trained with contextually consistent image-text pairs, fall short of addressing the human manipulation intention in CIR tasks. In this work, we introduce the construction of contextually complementary pseudo triplets from existing image-text pairs, unifying the pre-train and inference objectives of ZS-CIR.

\noindent \textbf{Mapping Image as One Word.} Several methods \cite{li2020oscar,zhang2021vinvl} represent image regions as word tokens via VLP models, which rely on object detector efficacy. However, ZR-CIR tasks extend the alignment ability beyond objects to scenes, styles, attributes, \textit{ect}. Our method addresses this issue by employing pseudo triplet data, which maps a pseudo reference image to a pseudo word token and combines it with the caption to align with the target image. PALAVRA \cite{10.1007/978-3-031-20044-1_32} proposes personalized image retrieval via cycle contrastive loss, requiring class-wise and caption annotations. In contrast, our model facilitates fine-grained image-to-word mapping without additional annotations. Other approaches \cite{Kumari_2023_CVPR,mokady2021clipcap,zhu2023visualize,tam2023simple} utilize a single word token to represent multiple images of the same object for text-to-image generation. Our model obviates the need for costly image grouping or supervised training.

\section{Methodology}

\subsection{Preliminary}

Given a reference image space \( \mathcal{I} \) and a text description space \( \mathcal{T} \), Composed Image Retrieval (CIR) involves a user text modifier \( T \in \mathcal{T} \) describing hypothetical semantic changes to a reference image \( I_{r} \in \mathcal{I} \), aiming to retrieve its closest realization \( I^{r}_{t} \in \mathcal{D} \) from an image database \( \mathcal{D} = \{I_i, \ldots, I_n\} \). This task can be formalized as a scoring function \( \Phi : \mathcal{I} \times \mathcal{T} \times \mathcal{D} \rightarrow \mathbb{R} \). In standard CIR, \( \Phi \) is typically learned through supervised training. However, Zero-Shot CIR (ZS-CIR) approaches \cite{Saito_2023_CVPR, baldrati2023zero, tang2023contexti2w} bypass this requirement by fine-tuning specific modules to map the reference image into an associated text representation. Specifically, these methods learn a mapping function \( f_{\theta}: \mathcal{I} \rightarrow \mathcal{Z} \), where \( \mathcal{Z} \) is a pre-defined text-token embedding space. This function \( f_{\theta} \) is trained using intermediate image representations from a specific image encoder \( \Psi_I \), often part of a pre-trained vision-language representation system like CLIP \cite{radford2021learning}. Template filling around the text modifier over the corresponding pseudo token embedding \( S_{*} = f_{\theta}(\Psi_I(I_{r})) \) is then employed to aggregate information into a target caption \( P \) (\textit{e.g.,} ``\texttt{a photo of \( S_{*} \), \( T \)).}'' This target caption serves as input for target image retrieval by vision-language models such as CLIP, encoding it using the associated pre-trained text encoder \( \Psi_T \) to project the target caption and candidate images \( I_i \in \mathcal{D} \) into a shared embedding space. The respective matching score is computed using cosine similarity \( \texttt{cos\_sim}(\Psi_I(I_{r}), \Psi_T(P)) \).

Figure \ref{fig:model-architecture} illustrates our approach. We first introduce the automatic \textit{Pseudo Triplet Construction} (PTC for short) process, leveraging the vision-language alignment knowledge from the pre-trained CLIP model \cite{radford2021learning} for constructing pseudo triplets with for object-compose and style-transfer. Subsequently, we introduce a novel \textit{Pseudo Composed Mapping} (PCM for short) approach to train ZS-CIR models with the pseudo-triplets with pseudo-manipulation description, addressing the discrepancy between the training and inference phases to filter intention-irrelevant visual information, achieving denoising image-to-word mapping. Our method is model-agnostic, and we incorporated three ZS-CIR models: Pic2Word \cite{Saito_2023_CVPR}, SEARLE \cite{baldrati2023zero}, LinCIR \cite{gu2024lincir}, and the SoTA Context-I2W \cite{tang2023contexti2w}. These models apply to a pre-trained mapping network to transform the information of reference image $I$ or manipulation text into a pseudo-word token $S_{*}$. To combine $I$ and $T$ across different modalities for zero-shot image retrieval, we form a sentence $P$ ``a photo of $S_{*}$ that {$T$}''. This query is then embedded using the frozen text encoder of CLIP. For each candidate image $I_c$, we generate embedding the frozen image encoder of CLIP. ZS-CIR is regarded as a traditional text-to-image retrieval task, where the similarity between $P$ and $I_c$ is evaluated.

\subsection{Automatic Pseudo Triplet Construction}

To reduce the task discrepancy between pre-training and retrieval in existing ZS-CIR models, which causes existing ZS-CIR models to only coarse alignment between the reference image and language tokens due to the absence of manipulative text descriptions. In existing ZS-CIR models, the reference image is indeed not aligned with the intentions of human manipulation, challenging the model's goal of efficiently intention-driven visual content extraction in the retrieval stage. We aim to construct triplets with pseudo-reference images and pseudo-manipulation text for training mapping networks to denoising intention-irrelevant visual information during mapping.  

Furthermore, to facilitate the seamless integration of PTC with existing ZS-CIR frameworks, we automatically generate pseudo triplets from existing image-caption pairs during the training stage.   Specifically, we leverage the image-caption pairs as target images and pseudo-manipulation text. Subsequently, we construct pseudo-reference images in \textit{partial-cropping} or \textit{style-transferring} the target images. These modifications ensure the pseudo-reference images contextually complement the pseudo-manipulation texts, thus enabling more accurate retrieval of target images. The process, illustrated in Figure \ref{fig:model-architecture}(left), contains the following three stages:

\begin{algorithm}
\caption{Pseudo Triplet Construction}
\label{alg:1}
\begin{algorithmic}[1]

\State \textbf{Input:} Batch of images $\mathcal{I}$, image captions $\mathcal{T}$
\State \textbf{Output:} Pseudo-triplets $\mathcal{P}_{tri}$

\Function{PTC}{$\mathcal{I}$, $\mathcal{T}$}
    \State Initialize $\mathcal{I}_c$ for cropped images
    \State Initialize $\mathcal{P}_{tri}$ for pseudo-triplets
    \State Compute embeddings $\boldsymbol{V}, \boldsymbol{V_c}, \boldsymbol{T}$ by CLIP

    \ForAll{$I_i$ in $\mathcal{I}$}
        \State $I_{ci} \gets \operatorname{Crop}(I_i, \text{random cropped size})$
        \State Add $I_{ci}$ to $\mathcal{I}_c$
    \EndFor

    \State Calculate $\text{Sim}_{T2V_c}, \text{Sim}_{T2V}$
    \For{$i=1$ to $\text{length}(\mathcal{I})$}
        \If{$\text{conditions for pseudo triple}$}
            \State Add $(\boldsymbol{v}_{ci}, T_i, \boldsymbol{v}_{i})$ to $\mathcal{P}_{tri}$
        \EndIf
    \EndFor

    \State Calculate $\text{Sim}_{V_c2V_c}, \text{Sim}_{V2V}$
    \ForAll{$\text{triples in } \mathcal{P}_{tri}$}
        \State $\boldsymbol{v}_{ri} \gets \text{Determine based on criteria}$
        \State Update triple with $\boldsymbol{v}_{ri}$
    \EndFor

    \State \textbf{return} $\mathcal{P}_{tri}$
\EndFunction

\end{algorithmic}
\end{algorithm}

\noindent \textbf{Partial Image Cropping for Object-level Mapping.} In this stage, we aim to construct pseudo-reference images that contextually complement pseudo-manipulation text at the object level(\textit{e.g., changing the number of objects or adding relationships with different objects}. This process involves cropping the target images. Specifically, for a batch of image $\mathcal{I}$ =$\{I_i\}_{i=1}^m$, where $m$ represents the batch size. Each image \( I_i \) in this batch undergoes a randomized cropping process to obtain a set of cropped images \( \mathcal{I}_c = \{I_{ci}\}_{i=1}^m \). For each \( I_i \), the cropping operation is defined by:
\begin{equation}
\begin{aligned}
I_{ci} = \operatorname{Crop}\left(I_i, (x_i, y_i, x_i + w_{ci}, y_i + h_{ci})\right)
\end{aligned}
\label{f:crop}
\end{equation}
where $\operatorname{Crop}(\cdot)$ denotes cropping operation, \( w_{ci} \) and \( h_{ci} \) are the width and height of the crop, and \( x_i \) and \( y_i \) are the top-left coordinates. These parameters are chosen with an element of randomness, introducing variability into the cropping process.  To ensure a partial overlap between the context of the cropped image and its caption, thereby enhancing the alignment between the images and their respective captions through contextually diverse. We limit the width and height cropping box to a size range of \(32\sim64\) pixels and position away from the image's central region.

\noindent \textbf{Cropped Image Filtering for Semantic Consistency.} The randomly cropped images frequently include potential noise irrelevant to the context required by the pseudo-manipulation text for composing retrieval target images. To filter out this noise and ensure the quality of the pseudo-triples, we employ the vision-language alignment knowledge from the pre-trained CLIP model. This enables us to select pseudo-reference images that adequately exclude irrelevant contexts and target images that sufficiently incorporate relevant contexts of the pseudo-manipulation texts. Specifically, we first utilize the visual encoder of the frozen CLIP to represent the original images and corresponding cropping images in a batch by image embeddings $\boldsymbol{V} = \{\boldsymbol{v}_i\}_{i=1}^m \subseteq \mathbb{R}^{d \times m}$ and $\boldsymbol{V_{c}} = \{\boldsymbol{v}_{ci}\}_{i=1}^m \subseteq \mathbb{R}^{d \times m}$, respectively. Then, the corresponding image captions are embedded by the language encoder of the frozen CLIP. We utilize the \texttt{[CLS]} token embeddings $\boldsymbol{T} = \{\boldsymbol{t}_i\}_{i=1}^m \subseteq \mathbb{R}^{d \times m}$ as the manipulation embeddings of the pseudo triplet. Where $d = 768$, and $m$ is the batch size. We calculate the similarity between each manipulation text \(\boldsymbol{t}_i\) and its composed cropped image \(\boldsymbol{v}_{ci}\) by \(\text{Sim}(\boldsymbol{t}_i, \boldsymbol{v}_{ci}) = \boldsymbol{t}_i{\boldsymbol{v}}^T_{ci}\) and its original image \(\boldsymbol{v}_{i}\) by \(\text{Sim}(\boldsymbol{t}_i, \boldsymbol{v}_{i}) = \boldsymbol{t}_i{\boldsymbol{v}}^T_{i}\). Finally, high-quality pseudo triplets $\mathcal{P}_{tri}$ are selected based on the criterion that negatively cropped images supplement the missing context in the caption, and positive target images ensure the quality of an image-caption pair as follows:

\begin{equation}
\begin{aligned}
\mathcal{P}_{tri} = \{ &(\boldsymbol{v}_{ci}, T_i, \boldsymbol{v}_{i}) \ | \ \text{Sim}(\boldsymbol{t}_i, \boldsymbol{v}_{ci}) < \Theta_{\boldsymbol{t}2\boldsymbol{v}_c}, \\
                         & \text{and} \ \text{Sim}(\boldsymbol{t}_i, \boldsymbol{v}_{i}) > \Theta_{\boldsymbol{t}2\boldsymbol{v}} \}
\end{aligned}
\label{f:select}
\end{equation}
where \( \Theta_{\boldsymbol{t}2\boldsymbol{v}_c} \) and \( \Theta_{\boldsymbol{t}2\boldsymbol{v}} \) are average for \( \text{Sim}(\boldsymbol{t}_i, \boldsymbol{v}_{ci}) \) and \( \text{Sim}(\boldsymbol{t}_i, \boldsymbol{v}_{i}) \) in a batch. $T_i$ donates the image caption as the manipulation text description. In this way, these pseudo triplets effectively simulate foreground/background differentiation and multi-object manipulation tasks.

\noindent \textbf{Image Style Transfer for Style-level Mapping.} The pseudo-reference images only contain cropped images that cannot imitate the style-level human manipulation intention (\textit{e.g.,} style transferring, attribute editing, and scene adjusting) in CIR tasks. To address this limitation, We mine semantically similar but visually distinct images as reference images, simulating style-level human manipulation intention. Specifically, we calculate the cropped and original images' similarity matrix \(\text{Sim}_{V_c2V_c} = \frac{\boldsymbol{V_c}}{\|\boldsymbol{V_c}\|} \cdot \frac{\boldsymbol{V_c}^T}{\|\boldsymbol{V_c}\|}\) and \(\text{Sim}_{V2V} = \frac{\boldsymbol{V}}{\|\boldsymbol{V}\|} \cdot \frac{\boldsymbol{V}^T}{\|\boldsymbol{V}\|}\), respectively. 
For the $i$-th sample in the pseudo triplets $\mathcal{P}{tri}$, we construct the reference image $\boldsymbol{v}_{ri}$ by a random composition: 65\% from the cropped images, 25\% from cropped of different images, and the 10\% from other original images (ablation in Table \ref{tab:more_ablation}) in a batch, as determined criteria:
\begin{equation}
\resizebox{0.908\hsize}{!}{$\begin{aligned}
\boldsymbol{v}_{ri} = 
     \begin{cases} 
     \boldsymbol{v}_{cj}, j= \text{argmax}(\{ \text{Sim}_{V_c2V_c}[i, j] \ | \ j \neq i \}) & \text{if } \boldsymbol{x} < 0.25 \\
     \boldsymbol{v}_{j}, j=\text{argmax}(\{ \text{Sim}_{V2V}[i, j] \ | \ j \neq i \}) & 0.25 \leq \boldsymbol{x} < 0.35   \\
     \boldsymbol{v}_{ci}, \text{otherwise}
     \end{cases}
\end{aligned}$}
\label{f:mining}
\end{equation}
where $\boldsymbol{x}$ is a randomly generated number within the range of $0\sim1$. This approach ensures that the context of the substitute image closely resembles the original reference image. Consequently, these pseudo triplets are effectively utilized to simulate color variations and domain conversion tasks.

\textbf{Algorithm for Pseudo Triple Construction}. The pseudo-code for our Pseudo Triple Construction (PTC) module is delineated in Algorithm \ref{alg:1}. We generate pseudo-triplets for object-level and style-level mapping from a batch of image-caption pairs. The process initiates with randomized cropping of each image in the batch, generating a set of cropped images denoted as \(I_{ci}\). These randomly cropped images often contain potential noise, which is irrelevant to the context required by the pseudo-manipulation text for composing retrieval target images. To mitigate this noise and enhance the quality of the pseudo-triples, we leverage the vision-language alignment capabilities of the pre-trained CLIP model. This model aids in selecting pseudo-reference images that effectively exclude irrelevant contexts and target images that aptly include the relevant contexts of the pseudo-manipulation texts.

Specifically, we utilize the visual encoder of the frozen CLIP to represent both the original and the cropped images within a batch through image embeddings, denoted as \(\boldsymbol{V}\) and \(\boldsymbol{V_c}\), respectively. The image captions are then embedded using the language encoder of the frozen CLIP, with the \texttt{[CLS]} token embeddings \(\boldsymbol{T}\) serving as the manipulation text embeddings of the pseudo triplet. We subsequently calculate the similarity between each manipulation text embedding and its corresponding cropped image embedding, \(\text{Sim}_{\boldsymbol{T2V_c}}\), and its original image, \(\text{Sim}_{\boldsymbol{T2V}}\). High-quality pseudo triplets \(\mathcal{P}_{tri}\) are then selected based on a criterion that ensures negatively cropped images supplement the missing context in the caption while positively targeted images maintain the integrity of an image-caption pair.

To refine the pseudo triplets' capabilities for style transfer and attribute editing, essential for style-level human manipulation intention mapping in CIR tasks, we implement a mining strategy to select visually similar yet distinct original images within the batch as pseudo-reference images. We first calculate the similarity matrices for the cropped and original images, \(\text{Sim}_{\boldsymbol{V_c2V_c}}\) and \(\text{Sim}_{\boldsymbol{V2V}}\), respectively. Original images are then chosen as pseudo-reference images based on a predefined criterion that ensures the context of the substitute image closely aligns with that of the original reference image. These carefully curated pseudo triplets are instrumental in simulating color variations and facilitating domain conversion tasks.

\subsection{Pseudo Composed Mapping}

The training framework in existing ZS-CIR models typically involves aligning the image with a pseudo-token \(S_{*}\) and composing a query using a simple prompt (\textit{e.g.,} ``\texttt{a photo of \(S_{*}\)}''). This approach aims to learn the equivariance transformation between the query and the original image. However, during the inference phase, the objective shifts to modifying reference images to retrieve target images, diverging from the principle of equivariance transformation. The absence of manipulation text and reference images in the training scheme complicates the selection of intention-relevant visual information for mapping, resulting in inaccurate retrieval. 

To address these limitations, we propose PCM, a novel training scheme that leverages our pseudo-triplets with pseudo-manipulation intention to learn intention-based image-to-word mapping. This revised training approach enhances the model's ability to adapt to the intent-special visual information required during the retrieval stage, denoising the heavy information redundancy in the pseudo token $S_*$ and significantly improving retrieval accuracy without necessitating complex modifications to the model architecture. Moreover, our training scheme maintains the existing model structure, thereby facilitating seamless integration into current ZS-CIR models. 

Figure \ref{fig:model-architecture} (upper right) provides a detailed illustration of our training scheme. We utilize pseudo-triplets to train an image-to-word mapping network, enabling it to adapt to the intent-specific visual information needed during the retrieval stage. To minimize redundant visual information and concentrate on the visual information relevant to manipulation intention, we introduce two innovative mapping processes, each accompanied by novel losses inspired by contrastive loss: (1) \textit{Composed Mapping Process}, maps the image to denoising word with intent-special visual information and reduce the task discrepancy between training and retrieval stages in existing ZS-CIR models. (2)  \textit{Global Mapping Process}, maps the image to word globally through equivariance transformation, enhancing the image-to-word mapping to balance the influence of pseudo-manipulation texts during training.

\noindent \textbf{Pseudo-Compose Mapping Process.} Given a pseudo triple with pseudo-manipulation intention, this process aims to train an image-to-word mapping network to denoising intention-irrelevant visual information. Specifically, For a pseudo triple that contains a pseudo-reference image embedding $\boldsymbol{v}_r$, a target image embedding $\boldsymbol{v}$, and a pseudo-manipulation text $T$. We first construct a prompt sentence ``\texttt{a photo of [*], \{T\}}'' with a manipulation intention $T$ to map and supplement the reference image, facilitating the retrieval of target images with comprehensive information. Then, we extract its pseudo-word token embedding $\boldsymbol{S}_{*} = f_M(\boldsymbol{v}_r)$. 

As shown in the Figure \ref{fig:model-architecture}, we replace $\boldsymbol{S}_{*}$ at \texttt{[*]} token embeddings of the fine-grained prompt sentence and feed it to the language encoder of CLIP to obtain the sentence embedding $\boldsymbol{t}_{s}$.  Finally, we propose a cross-modal compose loss $\mathcal{L}_{compose}$, which aims to match a target image embedding $\boldsymbol{v}$ to its prompt sentence embedding $\boldsymbol{t}_{s}$ while separating unpaired ones as follows:
\begin{equation}
\begin{aligned}
\mathcal{L}_{compose}=\mathcal{L}_{t 2 i}(\hat{\boldsymbol{t}}_{s},\hat{\boldsymbol{v}})+\mathcal{L}_{i 2 t}(\hat{\boldsymbol{t}}_{s}, \hat{\boldsymbol{v}})
\end{aligned}
\label{f:compose_loss}
\end{equation}

The two contrastive loss terms with a temperature hyper-parameter $\tau$ that controls the strength of penalties on hard negative samples are defined as:
\begin{equation}
\begin{aligned}
\mathcal{L}_{t 2 i}(\hat{\boldsymbol{t}}_{s}, \hat{\boldsymbol{v}})=-\frac{1}{|\mathcal{B}|} \sum_{i \in \mathcal{B}} \log \frac{\exp \left(\tau \hat{\boldsymbol{t}}_{s_{i}}^{T} \hat{\boldsymbol{v}}_{{i}}\right)}{\sum_{j \in \mathcal{B}} \exp \left(\tau \hat{\boldsymbol{t}}_{s_{i}}^{T} \hat{\boldsymbol{v}}_{{j}}\right)}
\end{aligned}
\label{f:CTL_1}
\end{equation}
\begin{equation}
\begin{aligned}
\mathcal{L}_{i 2 t}(\hat{\boldsymbol{t}}_{s}, \hat{\boldsymbol{v}})=-\frac{1}{|\mathcal{B}|} \sum_{i \in \mathcal{B}} \log \frac{\exp \left(\tau \hat{\boldsymbol{v}}_{{i}}^{T} \hat{\boldsymbol{t}}_{s_{i}}\right)}{\sum_{j \in \mathcal{B}} \exp \left(\tau \hat{\boldsymbol{v}}_{{i}}^{T} \hat{\boldsymbol{t}}_{s_{j}}\right)}
\end{aligned}
\label{f:CTL_2}
\end{equation}
\noindent where $\hat{\boldsymbol{t}}_{s_{i}} = \frac{\boldsymbol{t}_{s_{i}}}{\|\boldsymbol{t}_{s_{i}}\|}$ and $\hat{\boldsymbol{v}}_{{j}} = \frac{\boldsymbol{v}_{{j}}}{\|\boldsymbol{v}_{{j}}\|}$ are the normalized features of $i$-th prompt sentence embedding $\boldsymbol{t}_{s_i}$ and the $j$-th reference image embedding $\boldsymbol{v}_{{j}}$ in a batch $\mathcal{B}$. In this way, the discrepancy between training and inference in existing ZS-CIR models is reduced for mapping an image to a denoising pseudo-word token with intent-special visual information.

\noindent \textbf{Global Mapping Process.} During the training of the compose image process, the frozen CLIP language encoder initially tends to prioritize the pseudo-manipulation text, which it processes more readily than the simpler pseudo token \(S_*\). This preference can lead to model collapse, where the mapping network erroneously aligns various images to similar pseudo tokens. To address this, we introduce a global mapping process that omits pseudo-manipulation texts, thereby ensuring that the model focuses exclusively on image-to-word mapping.

Starting with a reference image embedding \(\boldsymbol{v}_r\), we map this image to a pseudo-word token \(\boldsymbol{S_{r}}_{*} = f_M(\boldsymbol{v}_r)\). As detailed in Figure \ref{fig:model-architecture}, we then append \(\boldsymbol{S_{r}}_{*}\) to the end of the token embeddings of the globally prompted sentence, ``\texttt{a photo of}'', and input this composition as a query into the language encoder of CLIP to derive the sentence embedding \(\boldsymbol{t}_{rs}\). Following this, we introduce a cross-modal alignment loss that aims to accurately match each reference image \(\boldsymbol{v}_r\) with its corresponding prompt sentence \(\boldsymbol{t}_{rs}\) while effectively distinguishing and separating unpaired images as follows:
\begin{equation}
\begin{aligned}
\mathcal{L}_{align}=\mathcal{L}_{t 2 i}(\hat{\boldsymbol{t}}_{rs},\hat{\boldsymbol{v}}_r)+\mathcal{L}_{i 2 t}(\hat{\boldsymbol{t}}_{rs}, \hat{\boldsymbol{v}}_r)
\end{aligned}
\label{f:align_loss}
\end{equation}
\begin{table}[ht]
  \centering 
     \caption{The number of images used for evaluation in each dataset.}
  \scalebox{1.0}{
  \begin{tabular}{c|c|c} 
  \toprule
  \cmidrule{1-3}
  Dataset & Query images & Candidate images\\ 
  \midrule
  ImageNet &10,000& 16,983\\
  COCO &4,766&4,766\\ 
  CIRR (test) &4,148&2,315\\
  Fashion (Dress) & 2,017 & 3,817\\
  Fashion (Shirt) &2,038&6,346\\
   Fashion (TopTee) & 1,961&5,373\\
  \bottomrule
  \end{tabular}}

  \label{tab:dataset_details}
\end{table}
\noindent where $\mathcal{L}_{t 2 i}$ and $\mathcal{L}_{i 2 t}$ are contrastive loss terms as Eq.\ref{f:CTL_1} and Eq.\ref{f:CTL_2}, $\hat{\boldsymbol{t}}_{rs} = \frac{\boldsymbol{t}_{rs}}{\|\boldsymbol{t}_{rs}\|}$ and $\hat{\boldsymbol{v}}_{r} = \frac{\boldsymbol{v}_{r}}{\|\boldsymbol{v}_{r}\|}$ are the normalized features of prompt sentence embedding $\boldsymbol{t}_{rs}$ and the target image embedding $\boldsymbol{v}$.  In this way, the image-to-word mapping is enhanced, offsetting the impact of the text's richer semantics.

The ﬁnal loss used to optimize $f_M$ is:
\begin{equation}
\begin{aligned}
\min _{M} \mathcal{L}=\mathcal{L}_{compose} + \mathcal{L}_{align}
\end{aligned}
\label{f:train_loss}
\end{equation}
\subsection{Inference with Denoise-I2W}
In the inference stage, we compose the reference image with the manipulation text as a query and compare the composed query with candidate images for retrieval. As shown in Figure \ref{fig:model-architecture} (bottom right), we compose the pseudo token embedding $\boldsymbol{S}_*$ of the image from the mapping network with the text description and feed it to the pre-trained language encoder of CLIP.  The result is embedded by the text encoder and compared to the visual features of candidate images.

Since our focus is on studying the intention-based denoising word mapping for ZS-CIR, we utilize the same prompt in the most recent works \cite{Saito_2023_CVPR,tang2023contexti2w} for a fair comparison. We show prompt examples for different ZS-CIR tasks. In all examples, \texttt{[*]} indicates the pseudo token from the mapping network: \textbf{(a) Domain conversion} aims to modify the domain of the reference image. The prompt is defined as \texttt{a [domain tag] of [*]}; \textbf{(b) Object composition} retrieves an image that contains an object in the reference image and other object tags. The prompt is in the format of \texttt{a photo of [*], [obj$_1$ tag] and [obj$_2$ tag], $\dots$, and [obj$_n$ tag]}; \textbf{(c) Sentence manipulation} modifies the reference image based on a sentence. We simply append the sentence with the special token as  \texttt{a photo of [*], [sentence]}.

\section{Experiments}

\noindent \textbf{Datasets.} We evaluate our model on four ZS-CIR datasets, \textit{i.e.,} COCO \cite{10.1007/978-3-319-10602-1_48} for object composition, ImageNet \cite{deng2009imagenet,Hendrycks_2021_ICCV} for domain conversion, CIRR \cite{Liu_2021_ICCV} for object/scene manipulation, and Fashion-IQ \cite{Wu_2021_CVPR} for attribute manipulation. All the dataset settings and evaluation metrics (Recall@K) follow the recent works \cite{Saito_2023_CVPR,tang2023contexti2w} for a fair comparison.   The evaluation datasets are preprocessed, as explained in the main paper. We describe the details of the dataset, \textit{i.e.,} number of query images and candidate images used for evaluation in Table \ref{tab:dataset_details}.

\noindent \textbf{(1) Domain conversion}. This setup evaluates the ability to compose real images and domain information to retrieve corresponding domain-specific images. We utilize ImageNet \cite{deng2009imagenet} and ImageNet-R \cite{Hendrycks_2021_ICCV}, which comprises 200 classes with diverse domains and has domain annotations. Following Pic2Word, we pick cartoons, origami, toys, and sculptures as the evaluation targets to avoid noise in the annotations. With this selection, we have 16,983 images as candidates. In the evaluation, given the real image from ImageNet and target domain names, we compose the query following the procedure in (a) in the Inference section. \textit{e.g.,} \texttt{a cartoon of [*]}.

\noindent \textbf{(2) Object composition}.  We evaluate the validation split (5000 images) of COCO \cite{10.1007/978-3-319-10602-1_48}, which dataset contains images with corresponding lists of object classes and instance mask of query images. Following Pic2Word, we randomly crop one object and mask its background using its instance mask to create a query for each image. The list of object classes is used as text specification. Given the reference image and class list, we compose a query by following (b) in the Inference section. \textit{e.g.,} \texttt{a photo of [*], [cat] and [dog]}.

\noindent \textbf{(3) Object/scene manipulation by text description}. In this setup, a reference image is provided alongside a text description containing instructions for manipulating either an object or the background scene depicted in the reference image. This composition of the reference image and text description enables the retrieval of manipulated images. We evaluate the test split of CIRR \cite{Liu_2021_ICCV} using the standard evaluation protocol following previous works \cite{Saito_2023_CVPR,baldrati2023zero,tang2023contexti2w}, and query texts are composed following the procedure in (c) of the Inference.

\noindent \textbf{(4) Attribute manipulation}. We employ Fashion-IQ \cite{Wu_2021_CVPR}, which includes various modification texts related to image attributes. These attribute manipulations are given as a sentence. As with CIRR, we adopt the standard evaluation protocol and create query texts following the procedure provided in (c) of the Inference section. In evaluation, we employ the validation set, following previous works \cite{Baldrati_2022_CVPR,Saito_2023_CVPR,baldrati2023zero,tang2023contexti2w}.

\begin{figure*}[t]
    \centering
    \centering
    \includegraphics[width=0.75\linewidth]{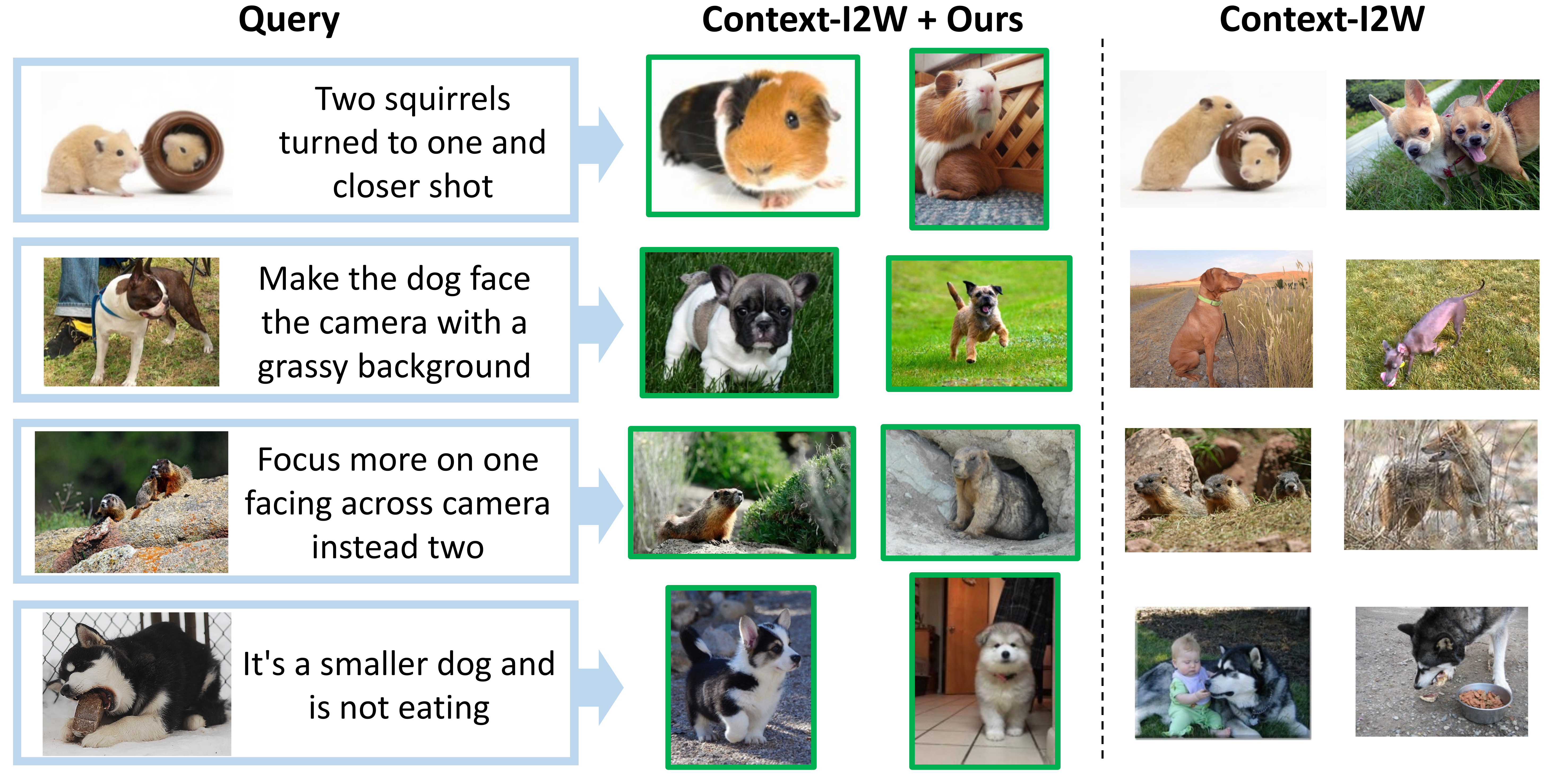}
    \caption{Visualization of retrieved results on the object manipulation task. Our Denosing-I2W enables ZS-CIR models to map intent-specific object/scene information, thereby enhancing the model’s ability to understand complex objects/scenes.}
    \label{fig:cirr}
\end{figure*}

\begin{table}[t]
\centering
\caption{Results on CIRR for object manipulation.}
\scalebox{1.0}
{\footnotesize
\begin{tabular}{cccccccccccc}
\toprule
Supervision                                      & Methods                                                   & R1                                               & R5                                               & \multicolumn{1}{l}{R10}                          & R50                                              \\ \midrule
\multicolumn{1}{c|}{}                            & \multicolumn{1}{c|}{Image-only}                           & 7.4                                              & 23.6                                             & 34.0                                             & 57.4                                             \\
\multicolumn{1}{c|}{}                            & \multicolumn{1}{c|}{text-only}                            & 20.9                                             & 44.8                                             & 55.5                                             & 79.1                                             \\
\multicolumn{1}{c|}{}                            & \multicolumn{1}{c|}{Image+Text}                           & 12.4                                             & 36.2                                             & 49.1                                             & 78.2                                             \\ 
\multicolumn{1}{c|}{}                            & \multicolumn{1}{c|}{LinCIR}                           & 25.0                                           & 53.3                                           & 66.7                                            & --                                            \\  \cmidrule(lr){2-6}
\multicolumn{1}{c|}{}                            & \multicolumn{1}{c|}{Pic2Word}                             & 23.9                                            & 51.7                                           & 65.3                                            & 87.8                                            \\
\multicolumn{1}{c|}{}                            & \multicolumn{1}{c|}{Pic2Word+our}                             & 24.7                                            & 53.2                                           & 66.9                                            & 88.1                                            \\ \cmidrule(lr){2-6}
\multicolumn{1}{c|}{}                            & \multicolumn{1}{c|}{SEARLE-XL}                   & 24.2                                            & 52.4                                            & 66.3                                            & 88.6                                   \\
\multicolumn{1}{c|}{}                            & \multicolumn{1}{c|}{SEARLE-XL+our}                   & 25.0                                            & 54.1                                            & 67.3                                            & 89.2                                   \\ \cmidrule(lr){2-6}
\multicolumn{1}{c|}{}                            & \multicolumn{1}{c|}{Context-I2W}                   & 25.6                                            & 55.1                                            & 68.5                                            & 89.8                                   \\
\multicolumn{1}{c|}{}                            & \multicolumn{1}{c|}{Context-I2W+our(50\%)}                         & 25.9                                   & 55.3                                   & 68.6                                   & 90.0                                   \\
\multicolumn{1}{c|}{\multirow{-11}{*}{ZERO-SHOT}}                             & \multicolumn{1}{c|}{\textbf{Context-I2W+our(100\%)}}                         & \textbf{26.9}                                   & \textbf{57.2}                                   & \textbf{69.8}                                   & \textbf{90.6}                                   \\
\cmidrule(lr){1-6}
\multicolumn{1}{c|}{CIRR}                        & \multicolumn{1}{c|}{Combiner}                             & 30.3                                             & 60.4                                             & 73.2                                             & 92.6                                             \\
\multicolumn{1}{c|}{Fashion-IQ}                  & \multicolumn{1}{c|}{Combiner}                             & 20.1                                             & 47.7                                             & 61.6                                             & 85.9                                             \\ 
\multicolumn{1}{c|}{CIRR}                        & \multicolumn{1}{c|}{ARTEMIS}                              & 17.0                                             & 46.1                                             & 61.3                                             & 87.7                                             \\
\multicolumn{1}{c|}{CIRR}                        & \multicolumn{1}{c|}{CIRPLANT}                             & 19.6                                             & 52.6                                             & 68.4                                             & 92.4                                             \\ \bottomrule
\end{tabular}}

\label{tab:CIRR}
\end{table}

\noindent \textbf{Implementation Details.}  We adopt ViT-L/14 CLIP \cite{radford2021learning} pre-trained on 400M image-text paired data. For training Pic2Word and Context-I2W, we utilize the Conceptual Caption dataset \cite{DBLP:conf/acl/SoricutDSG18}, which comprises 3M images in 20 hours and 28 hours, respectively.  We employ AdamW \cite{loshchilov2018decoupled} with a learning rate of $1\times10^{-5}$, weight decay of $0.1$, and a linear warmup of $10000$ steps. The batch size for contrastive learning is $1024$.For training SEARLE, we utilize the ImageNet1K \cite{deng2009imagenet} test set, which comprises 100K images in 1.5 hours, and leverage the image descriptions generated by SEARLE's GPT as captions. We employ AdamW, with a learning rate of $5\times10^{-5}$ and a batch size of $256$, the loss weights $L_{cos}$ and $L_{gpt}$ to 1 and 0.5, respectively. We adopt an exponential moving average with 0.99 decay. All models are trained on $2$ NVIDIA A100 (80G) GPUs and remain in the same setting mentioned in their papers. To ensure reliable results, we report the average performance over three trials.  Moreover, we conduct ablation studies on CIRR test sets and FashionIQ validation sets. For FashionIQ, we consider the average recall.

\subsection{Quantitative and Qualitative Results}
We integrate Denoise-I2W with several ZS-CIR methods, maintaining consistent training settings across these models, including:  1) \textbf{Pic2Word} \cite{Saito_2023_CVPR}: maps the entire visual features of a reference image into a pseudo-word token within the CLIP token embedding space. 2) \textbf{SEARLE-XL} \cite{baldrati2023zero}: Similar to Pic2Word, it further integrates the pseudo-word token with the relative caption generated by GPT \cite{brown2020language} and distill for efficiency. 3) \textbf{Context-I2W} \cite{tang2023contexti2w}: Selectively extracts text description-relevant visual information from the reference image before mapping it into a pseudo-word token. 4) \textbf{LinCIR} \cite{gu2024lincir}: Masks subjects in captions from various image-text datasets for language-only efficiency training. For a fair comparison, we present the reported results of methods relying on the ViT-L/14 CLIP model and do not compare those leveraging LLMs during inference  \cite{training-free-cir} due to their inefficiency. We evaluated Denoise-I2W's performance by comparing the reported results of these models. Moreover, we compare Denoise-I2W with the standard ZS-CIR method, standard ZS-CIR methods, including 5) \textbf{Text-only}: Computes similarity based on the CLIP features of descriptions and candidate images; 6) \textbf{Image-only}: Retrieves the most similar images to the reference image; and 7) \textbf{Image + Text}: Sums the CLIP features of the reference image and the description, and the published results of widely compared supervised models, including Combiner \cite{Baldrati_2022_CVPR}, ARTEMIS \cite{delmas2022artemis}, CIRPLANT \cite{Liu_2021_ICCV}.

\begin{table*}[t]
\centering
\caption{Results on Fashion-IQ for attribute manipulation.}
\scalebox{1.0}
{\footnotesize
\setlength{\tabcolsep}{3.5mm}
\begin{tabular}{cccccccccccc}
\toprule
                            &                                                          & \multicolumn{2}{c}{Dress}                                                               & \multicolumn{2}{c}{Shrit}                                                      & \multicolumn{2}{c}{TopTee}                                                              & \multicolumn{2}{c}{Average}                               \\ \cmidrule(lr){3-4}\cmidrule(lr){5-6}\cmidrule(lr){7-8}\cmidrule(lr){9-10}
Supervision                 & Methods                                                  & R10                                  & R50                                              & R10                         & R50                                              & R10                                  & R50                                              & R10                         & R50                         \\ \cmidrule(lr){1-10}
                           & Image-only                                               & 5.4                                  & \multicolumn{1}{c|}{13.9}                        & 9.9                         & \multicolumn{1}{c|}{20.8}                        & 8.3                                  & \multicolumn{1}{c|}{17.7}                        & 7.9                         & 17.5                        \\
                           & Text-only                                               & 13.6                                 & \multicolumn{1}{c|}{29.7}                        & 18.9                        & \multicolumn{1}{c|}{31.8}                        & 19.3                                 & \multicolumn{1}{c|}{37.0}                        & 17.3                        & 32.9                        \\
                            & Image+Text                                               & 16.3                                 & \multicolumn{1}{c|}{33.6}                        & 21.0                        & \multicolumn{1}{c|}{34.5}                        & 22.2                                 & \multicolumn{1}{c|}{39.0}                        & 19.8                        & 35.7                        \\
                            & LinCIR (CVPR 2024)                                               & 20.9                                 & \multicolumn{1}{c|}{42.4}                        & 29.1                        & \multicolumn{1}{c|}{46.8}                        & 28.8                                 & \multicolumn{1}{c|}{50.2}                        & 26.3                        & 46.5                        \\ \cmidrule(lr){2-10}
                            & Pic2Word (CVPR 2023)                                                & 20.0                                 & \multicolumn{1}{c|}{40.2}                        & 26.2                        & \multicolumn{1}{c|}{43.6}                        & 27.9                                 & \multicolumn{1}{c|}{47.4}                        & 24.7                        & 43.7                        \\
                            & Pic2Word+Denoise-I2W                                                   & 21.6                                 & \multicolumn{1}{c|}{41.4}                        & 27.8                        & \multicolumn{1}{c|}{45.8}                        & 29.2                                 & \multicolumn{1}{c|}{50.0}                        & 26.2                        & 45.7                        \\ \cmidrule(lr){2-10}
                            & SEARLE-XL (ICCV 2023)                                               & 20.3                                 & \multicolumn{1}{c|}{43.2}                        & 27.4                                 & \multicolumn{1}{c|}{45.7}                        & 29.3                                 & \multicolumn{1}{c|}{50.2}                        & 25.7                        & 46.3                        \\
                            & SEARLE-XL+Denoise-I2W                                               & 22.7                                 & \multicolumn{1}{c|}{45.0}                        & 29.4                                 & \multicolumn{1}{c|}{47.9}                        & 30.2                                 & \multicolumn{1}{c|}{51.4}                        & 27.4                        & 48.1                        \\ \cmidrule(lr){2-10}
                            & Context-I2W (AAAI 2024)                                               & 23.1                                 & \multicolumn{1}{c|}{45.3}                        & 29.7                                 & \multicolumn{1}{c|}{48.6}                        & 30.6                                 & \multicolumn{1}{c|}{52.9}                        & 27.8                        & 48.9                        \\
                            & Context-I2W+Denoise-I2W(50\%)                                             & 23.3                        & \multicolumn{1}{c|}{45.7}               & 30.0                        & \multicolumn{1}{c|}{48.9}               & 31.0                        & \multicolumn{1}{c|}{53.2}               & 28.1               & 49.3               \\ 
                            \multirow{-11}{*}{ZERO-SHOT} & \textbf{Context-I2W+Denoise-I2W(100\%)}                                             & \textbf{24.4}                        & \multicolumn{1}{c|}{\textbf{47.8}}               & \textbf{30.9}                        & \multicolumn{1}{c|}{\textbf{49.8}}               & \textbf{31.6}                        & \multicolumn{1}{c|}{\textbf{54.1}}               & \textbf{29.0}               & \textbf{50.6}               \\ \cmidrule(lr){1-10}
CIRR                        & Combiner (CVPR 2022)                                                & 17.2                                 & \multicolumn{1}{c|}{37.9}                        & 23.7                        & \multicolumn{1}{c|}{39.4}                        & 24.1                                 & \multicolumn{1}{c|}{43.9}                        & 21.7                        & 40.4                        \\
Fashion-IQ                  & Combiner (CVPR 2022)                                                & 30.3                                 & \multicolumn{1}{c|}{54.5}                        & 37.2                        & \multicolumn{1}{c|}{55.8}                        & 39.2                                 & \multicolumn{1}{c|}{61.3}                        & 35.6                        & 57.2                        \\ 
Fashion-IQ                  & CIRPLANT (ICCV 2021)                                              & 17.5                                 & \multicolumn{1}{c|}{40.4}                        & 17.5                        & \multicolumn{1}{c|}{38.8}                        & 21.6                                 & \multicolumn{1}{c|}{45.4}                        & 18.9                        & 41.5                        \\
Fashion-IQ                  & ARTEMIS  (ICLR 2022)                                                & 27.2                                 & \multicolumn{1}{c|}{52.4}                        & 21.8                        & \multicolumn{1}{c|}{43.6}                        & 29.2                                 & \multicolumn{1}{c|}{54.8}                        & 26.1                        & 50.3                        \\
\bottomrule
\end{tabular}}
\label{tab:fashion}
\end{table*}

\begin{figure}[t]
    \centering
    \setlength{\belowcaptionskip}{-8pt}
    \includegraphics[width=1.0\linewidth]{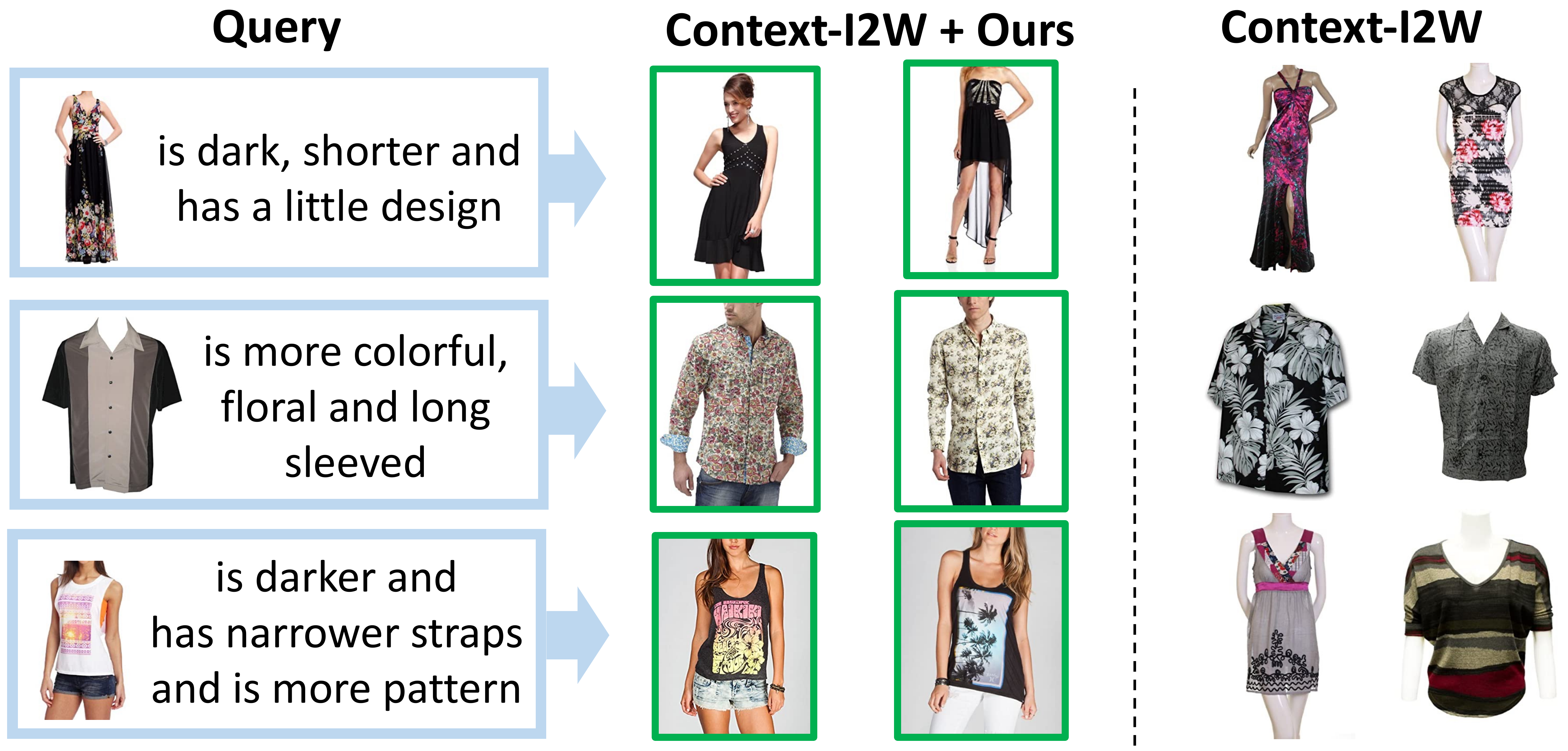}
    \caption{Visualization of retrieved results of attribute manipulation.}
    \label{fig:fashion}
\end{figure}

\begin{figure}[t]
    \centering
    \centering
    \includegraphics[width=1.0\linewidth]{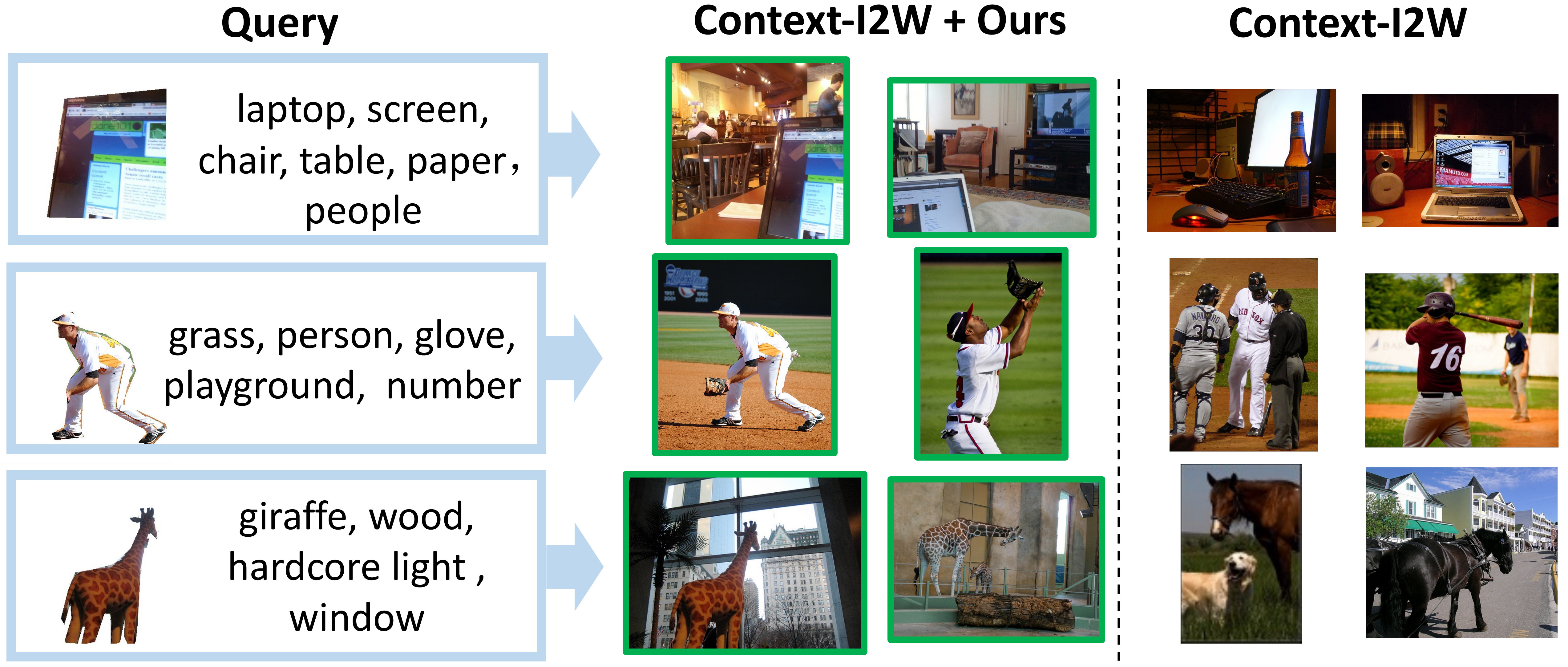}
    \caption{Retrieved results on the object composition task.}
    \label{fig:coco}
\end{figure}

\begin{figure}[t]
    \centering
    \centering
    \includegraphics[width=1.0\linewidth]{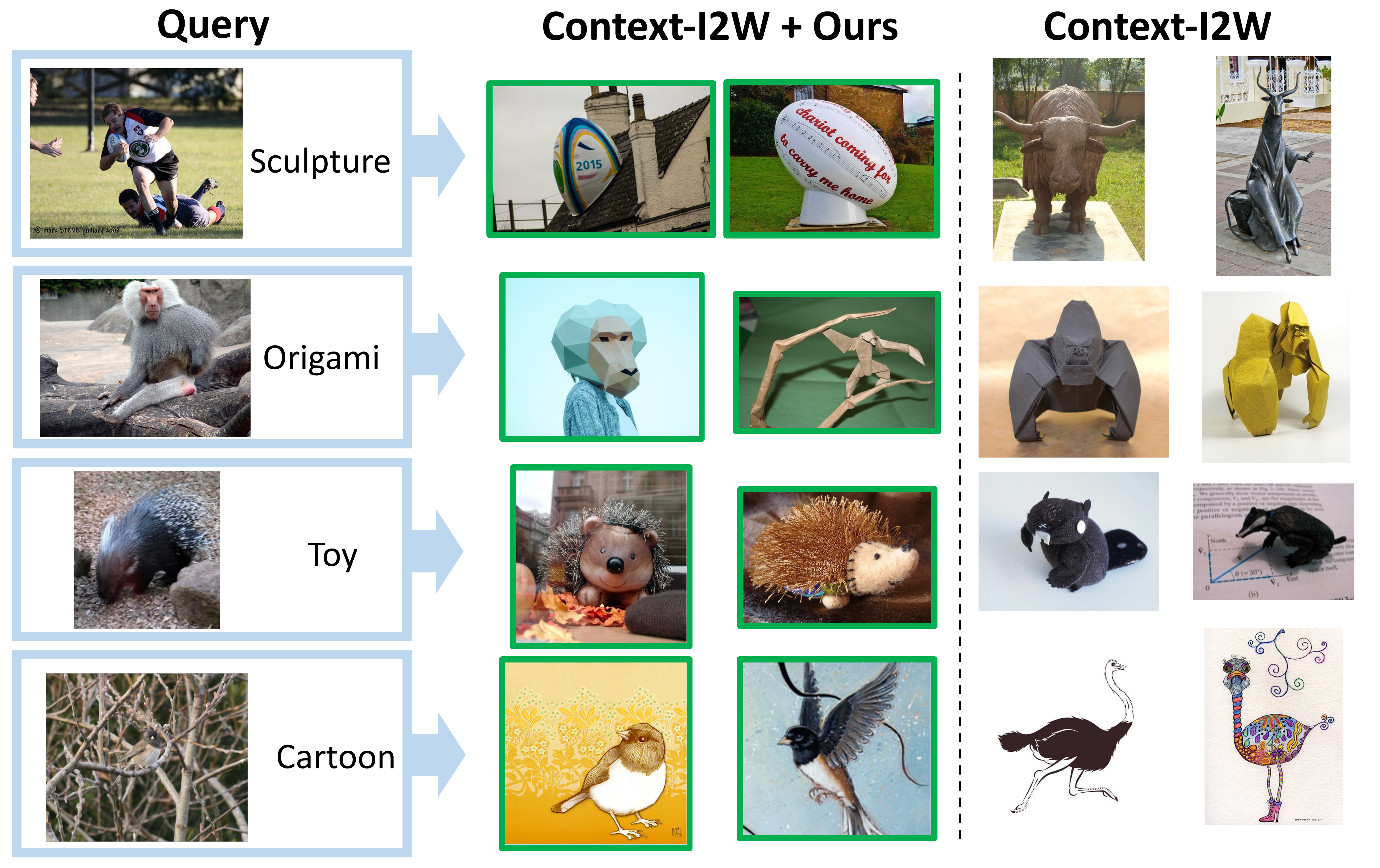}
    \caption{Retrieved results on the domain conversion task.}
    \label{fig:imgnet}
\end{figure}

Tables \ref{tab:fashion} to \ref{tab:coco} present the quantitative results, while Figures \ref{fig:cirr} to \ref{fig:imgnet} illustrate the corresponding qualitative results of our model alongside the latest Context-I2W. The integration of our Denoise-I2W with existing ZS-CIR methods markedly enhances performance across these methods. Specifically, in the object/scene manipulation task (Table \ref{tab:CIRR}), our approach demonstrates superior capabilities in fine-grained image editing and foreground/background differentiation. By incorporating Denoise-I2W with the SoTA Context-I2W, we consistently outperform existing models, achieving an average performance boost of 1.38\% over the SoTA model. This improvement is attributed to Denoise-I2W's PCM module to enable ZS-CIR models to map intent-specific object/scene information using pseudo triplets with partial-cropped images, thereby enhancing the model’s proficiency in understanding complex objects/scenes and significantly mitigating the impact of dataset-specific biases \cite{Saito_2023_CVPR}. As demonstrated in Figure \ref{fig:cirr}, Denoise-I2W enables the ZS-CIR model to accurately edit parallel objects (row 1), alter backgrounds (row 2), focus a special object (row 3), and even modify the state of one object while removing another overlapping object (row 4).

We further evaluate Denoise-I2W in attribute manipulation tasks that necessitate accurate localization of specific attributes across the entire image, guided by human manipulation text. As evidenced in Table \ref{tab:fashion}, integrating Denoise-I2W enables the SoTA model to achieve a notable average improvement of 1.45\%. This enhancement is attributed to Denoise-I2W's capability to enable ZS-CIR models to map attribute-relevant visual information effectively using pseudo triplets. These triplets incorporate varied image styles, thus enhancing the model's ability to identify specific attributes. Figure \ref{fig:fashion} further proves that Denoise-I2W separates visual details and selects fine-grained attributes, facilitating intention-based denoising word mapping.

In the domain conversion assessments (Table \ref{tab:imgnet}), Denoise-I2W, when integrated with the SoTA Context-I2W, consistently surpasses existing methods, including supervised approaches, with an average improvement of 2.20\%. Figure \ref{fig:imgnet} illustrates that Denoise-I2W enables ZS-CIR models to adeptly identify objects within complex scenes, such as a rugby ball in hand, a monkey on a rock, a hedgehog in a forest, and a sea lion in water. In contrast, Context-I2W struggles to accurately select the appropriate local visual features based on human manipulation text because it relies on a mapping from image-caption pairs devoid of manipulation intention, which is effectively addressed by our PTC module.

Furthermore, in the object composition tests (Table \ref{tab:coco}), Denoise-I2W enables the SoTA model to achieve a significant improvement of 4.17\% on average, outperforming supervised methods across all metrics. These results confirm the efficacy of our approach in precisely mapping visual information to the language token space. By bridging the gap between the training and retrieval phases, Denoise-I2W facilitates the combination of multiple objects, as evidenced in Figure \ref{fig:coco}. This integration enhances the ZS-CIR model's ability to handle complex image manipulations effectively.

\begin{table*}[t]
\centering
\caption{Results on ImageNet for domain conversion.}
\scalebox{1.0}
{\footnotesize
\setlength{\tabcolsep}{2.4mm}
\begin{tabular}{cccccccccccc}
\toprule
                           &              & \multicolumn{2}{c}{Cartoon}                       & \multicolumn{2}{c}{Origami}                        & \multicolumn{2}{c}{Toy}                            & \multicolumn{2}{c}{Sculpture}                      & \multicolumn{2}{c}{Average}   \\  \cmidrule(lr){3-4}\cmidrule(lr){5-6}\cmidrule(lr){7-8}\cmidrule(lr){9-10}\cmidrule(lr){11-12}
Supervision                & Methods      & R10          & R50                                & R10           & R50                                & R10           & R50                                & R10           & R50                                & R10           & R50           \\ \cmidrule(lr){1-12}
\multirow{9}{*}{ZERO-SHOT} & Image-only   & 0.3          & \multicolumn{1}{c|}{4.5}           & 0.2           & \multicolumn{1}{c|}{1.8}           & 0.6           & \multicolumn{1}{c|}{5.7}           & 0.3           & \multicolumn{1}{c|}{4.0}           & 0.4           & 4.0           \\
                           & Text-only   & 2.2          & \multicolumn{1}{c|}{1.1}           & 0.8           & \multicolumn{1}{c|}{3.7}           & 0.8           & \multicolumn{1}{c|}{2.4}           & 0.4           & \multicolumn{1}{c|}{2.0}           & 0.5           & 2.3           \\
                           & Image+Text   & 2.2          & \multicolumn{1}{c|}{13.3}          & 2.0           & \multicolumn{1}{c|}{10.3}          & 1.2           & \multicolumn{1}{c|}{9.7}           & 1.6           & \multicolumn{1}{c|}{11.6}          & 1.7           & 11.2          \\ \cmidrule(lr){2-12}
                           & Pic2Word (CVPR 2023)    & 8.0          & \multicolumn{1}{c|}{21.9}          & 13.5          & \multicolumn{1}{c|}{25.6}          & 8.7           & \multicolumn{1}{c|}{21.6}          & 10.0          & \multicolumn{1}{c|}{23.8}          & 10.1          & 23.2          \\ 
                           & Pic2Word+Denoise-I2W    & 9.2          & \multicolumn{1}{c|}{24.7}          & 15.8          & \multicolumn{1}{c|}{27.9}          & 10.2           & \multicolumn{1}{c|}{25.7}          & 11.6          & \multicolumn{1}{c|}{25.2}          & 11.7          & 25.9          \\  \cmidrule(lr){2-12}
                           & Context-I2W (AAAI 2024) & 10.2 & \multicolumn{1}{c|}{26.1} & 17.5 & \multicolumn{1}{c|}{28.7} & 11.6 & \multicolumn{1}{c|}{27.4} & 12.1 & \multicolumn{1}{c|}{28.2} & 12.9 & 27.6 \\ 
                           & Context-I2W+Denoise-I2W(50\%) & 10.1 & \multicolumn{1}{c|}{26.4} & 17.8 & \multicolumn{1}{c|}{30.2} & 12.1 & \multicolumn{1}{c|}{28.7} & 12.3 & \multicolumn{1}{c|}{29.2} & 13.2 & 28.6 \\
                           & \textbf{Context-I2W+Denoise-I2W(100\%)} & \textbf{10.7} & \multicolumn{1}{c|}{\textbf{26.9}} & \textbf{19.2} & \multicolumn{1}{c|}{\textbf{33.8}} & \textbf{13.5} & \multicolumn{1}{c|}{\textbf{30.2}} & \textbf{13.7} & \multicolumn{1}{c|}{\textbf{31.4}} & \textbf{14.3} & \textbf{30.6} \\ \cmidrule(lr){1-12}
CIRR                       & Combiner (CVPR 2022)    & 6.1          & \multicolumn{1}{c|}{14.8}          & 10.5          & \multicolumn{1}{c|}{21.3}          & 7.0           & \multicolumn{1}{c|}{17.7}          & 8.5           & \multicolumn{1}{c|}{20.4}          & 8.0           & 18.5          \\
Fashion-IQ                 & Combiner (CVPR 2022)    & 6.0          & \multicolumn{1}{c|}{16.9}          & 7.6           & \multicolumn{1}{c|}{20.2}          & 2.7           & \multicolumn{1}{c|}{10.9}          & 8.0           & \multicolumn{1}{c|}{21.6}          & 6.0           & 17.4          \\ \bottomrule
\end{tabular}}
\label{tab:imgnet}
\end{table*}

\begin{table}[t]
\centering
\caption{Results of the object composition task using COCO.}
\scalebox{1.0}
{\footnotesize
\begin{tabular}{cccccccccccc}
\toprule
Supervision                                     & Methods                           & R1            & R5            & \multicolumn{1}{l}{R10} \\  \midrule
\multicolumn{1}{c|}{\multirow{9}{*}{ZERO-SHOT}} & \multicolumn{1}{c|}{Image-only}   & 8.6           & 15.4          & 18.9                    \\
\multicolumn{1}{c|}{}                           & \multicolumn{1}{c|}{Text-only}   & 6.1           & 15.7          & 23.5                    \\
\multicolumn{1}{c|}{}                           & \multicolumn{1}{c|}{Image+Text}   & 10.2          & 20.2          & 26.6                    \\ \cmidrule(lr){2-5}
\multicolumn{1}{c|}{}                           & \multicolumn{1}{c|}{Pic2Word}     & 11.5          & 24.8          & 33.4                    \\
\multicolumn{1}{c|}{}                           & \multicolumn{1}{c|}{Pic2Word+our}     & 13.2          & 27.7          & 37.4                    \\ \cmidrule(lr){2-5}
\multicolumn{1}{c|}{}                           & \multicolumn{1}{c|}{Context-I2W}     & 13.5          & 28.5          & 38.1                    \\
\multicolumn{1}{c|}{}                           & \multicolumn{1}{c|}{Context-I2W+our(50\%)} & 14.5          & 29.9          & 39.4          \\
\multicolumn{1}{c|}{}                           & \multicolumn{1}{c|}{\textbf{Context-I2W+our(100\%)}} & \textbf{16.1}          & \textbf{33.0}          & \textbf{43.5}           \\ \cmidrule(lr){1-5}
\multicolumn{1}{c|}{CIRR}                       & \multicolumn{1}{c|}{Combiner}     & 9.9           & 22.8          & 32.2                    \\
\multicolumn{1}{c|}{Fashion-IQ}                 & \multicolumn{1}{c|}{Combiner}     & 13.2          & 27.1          & 35.2                   \\ \bottomrule
\end{tabular}}
\label{tab:coco}
\end{table}

\begin{figure*}[ht]
    \centering
    \centering
    \includegraphics[width=0.75\linewidth]{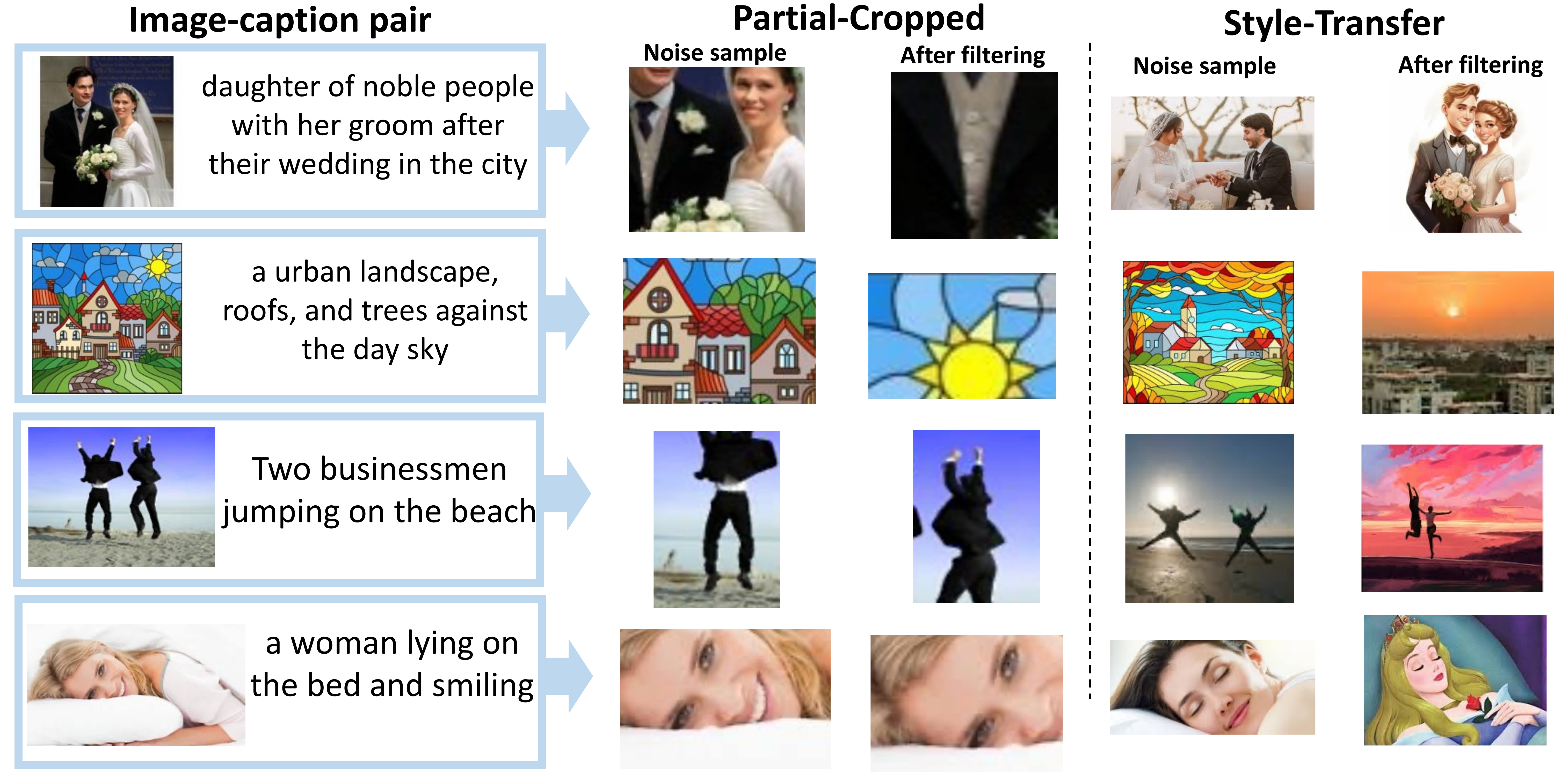}
    \caption{Example of noise samples and filter samples. Our filtering method effectively removes potential noise from pseudo triplets}
    \label{fig:noise}
\end{figure*}

\begin{table}[t]
\centering
\caption{Ablation study of main components on CIRR and FashionIQ.}
\scalebox{1.0}
{\footnotesize
\begin{tabular}{llcclcc}
\toprule
\multicolumn{2}{l}{}   & \multicolumn{3}{c}{CIRR}                         & \multicolumn{2}{c}{Fashion-IQ} \\ \cmidrule(lr){3-5}\cmidrule(lr){6-7}
   & Methods           & R1   & R5   & R10                                & R10            & R50           \\ \cmidrule(lr){1-7}
1. & full model        & 26.9 & 57.2 & \multicolumn{1}{l|}{69.8}          & 29.0           & 50.6         \\
2. & w/o cropped images       & 24.7 & 54.5 & \multicolumn{1}{l|}{68.0}          & 27.4           & 48.4          \\
3. & w/o mining images & 26.1 & 56.4 & \multicolumn{1}{l|}{69.3}          & 25.6           & 46.2          \\
4. & w/o select cropped images             & 24.0 & 53.5 & \multicolumn{1}{l|}{67.2}          & 25.7           & 46.1          \\
5. & w/o select target images       & 23.3 & 51.5 & \multicolumn{1}{c|}{65.7}          & 26.1           & 44.8          \\
6. & w/o compose loss          & 24.8 & 54.7 & \multicolumn{1}{c|}{68.2}          & 27.2           & 48.5          \\ 
7. & w/o alignment loss   & 23.5 & 53.8 & \multicolumn{1}{l|}{66.9}          & 25.7           & 46.8          \\
8. & mask images & 22.7 & 50.2 & \multicolumn{1}{l|}{64.5}          & 23.4           & 42.2          \\ 
9. & copped larger images  & 23.2 & 52.1 & \multicolumn{1}{c|}{65.2}          & 23.8           & 42.7          \\
\bottomrule
\end{tabular}}
\label{tab:ablation}
\end{table}

\begin{figure}[ht]
    \centering
    \includegraphics[width=1.0\linewidth]{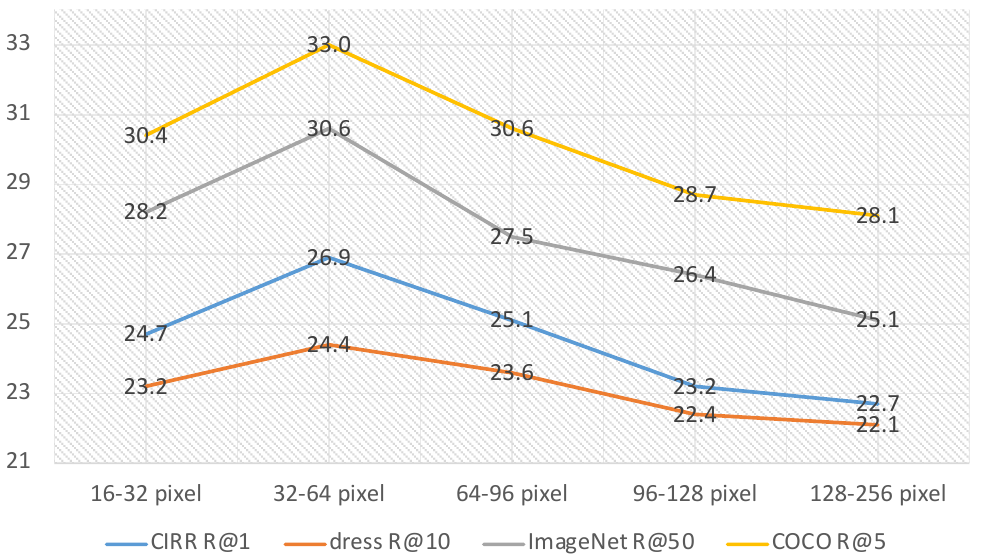}
    \caption{Analysis of the influence of cropped image's size.}
    \label{fig:crop_ablation}
\end{figure}

\begin{table}[ht]
\centering
\caption{Ablation study of key parameter selection on CIRR and FashionIQ.}
\scalebox{0.9}
{\footnotesize
\begin{tabular}{llcclcc}
\toprule
\multicolumn{2}{l}{}   & \multicolumn{3}{c}{CIRR}                         & \multicolumn{2}{c}{Fashion-IQ} \\ \cmidrule(lr){3-5}\cmidrule(lr){6-7}
   & Methods           & R1   & R5   & R10                                & R10            & R50           \\ \cmidrule(lr){1-7}
1. & 100\% cropped images & 24.9 & 55.4 & \multicolumn{1}{l|}{67.5}          & 27.5           & 48.5         \\ 
2. & 100\% original images & 24.1 & 54.6 & \multicolumn{1}{c|}{66.2}          & 26.9           & 48.1          \\
3. & 25\% cropped, 10\% original images       & 26.9 & 57.2 & \multicolumn{1}{l|}{69.8}          & 29.0           & 50.6         \\
4. & 25\% cropped, 25\% original images & 25.7 & 56.6 & \multicolumn{1}{l|}{68.6}          & 28.0           & 49.4         \\
5. & 10\% cropped, 10\% original images & 25.3 & 56.2 & \multicolumn{1}{l|}{67.5}          & 27.8           & 49.2         \\
6. & select relevant cropped images             & 23.4 & 52.2 & \multicolumn{1}{l|}{66.7}          & 24.3           & 43.3          \\
7. & select irrelevant target images       & 23.2 & 51.4 & \multicolumn{1}{c|}{65.1}          & 24.0           & 42.9          \\
\bottomrule
\end{tabular}}
\label{tab:more_ablation}
\end{table}

\subsection{Ablation Study}
In Table \ref{tab:ablation}, we appraise the influence of principal components in Denoise-I2W on the CIRR and FashionIQ datasets, employing the SoTA model, Context-I2W. \textbf{(1) Models `2-3' examine to evaluate the efficacy of the `partial-cropped' and `style-transfer' methods.} Without cropped images in model `2' leads to an average decrease of 2.23\% in CIRR, underscoring the importance of multi-object composition. Model `3', devoid of the replace mining images strategy, exhibits an average decline of 3.90\% in FashionIQ. This decline highlights the necessity for style-transfer images of attribute manipulation. \textbf{(2) Models `4-5' analysis of the role of image selection in alignment with caption context.} Ignoring to choose cropped images that complement the captions (Model `4') or target images that are relevant to the captions' context (Model `5') results in a marked performance decrease. \textbf{(3) Models `6-7' assess the significance of key modules within the PCM module.} Removing the composed loss in model `6' limits the model's ability to denoise irrelevant details in image-to-word mapping, causing a 2.02\% average decline. Without the alignment loss in model `7', the training of image-to-word mapping is highly influenced by the rich text descriptions, reducing average performance by 3.36\%. \textbf{(4) In models `8-9', we investigate alternative approaches to handling cropped images.} In model `8', we use randomly masked images, lacking context, resulting in a sharp 6.10\% decline, proving the effectiveness of Denoise-I2W's cropping method. Meanwhile, In model `9', a broader cropping range of $96 \sim 128$ pixel demonstrates a 5.30\% average reduction, approving the necessity of smaller-sized cropped images.

Table \ref{tab:more_ablation} presents ablation studies on key parameters within the Denoise-I2W model. \textbf{In models `1-2', assess the effectiveness of constructing reference images through random composition.} Using only cropped images from various sources (model `1') and different original images (model `2') led to performance declines of 1.94\% and 2.72\%, respectively, compared to our optimal configuration (model `3'). \textbf{In models `3-5', we evaluate the impact of adjusting the ratio of cropped to original images.} Increasing the proportion of original images (model `4') or reducing the proportion of cropped images (model `5') resulted in performance drops of 1.04\% and 1.50\%, respectively, highlighting the balance achieved in model `3', which utilizes 25\% cropped images and 10\% original images. \textbf{Models `6-7' examine the selection of images in our 'Cropped Image Filtering' approach.} Selecting contextually relevant cropped images (model `6') caused a significant performance decrease of 4.72\%, as the high context overlap complicates the model's learning of the denoting mapping. Conversely, using contextually irrelevant target images (model `7') resulted in an even greater average performance drop of 5.38\%, as it distorted the alignment between the text descriptions and the target images' context, leading to inaccuracies in the intention-based denoising image-to-word mapping. This emphasizes the precision required in image selection to maintain model efficacy.

\begin{figure*}[t]
    \centering
    \includegraphics[width=0.75\linewidth]{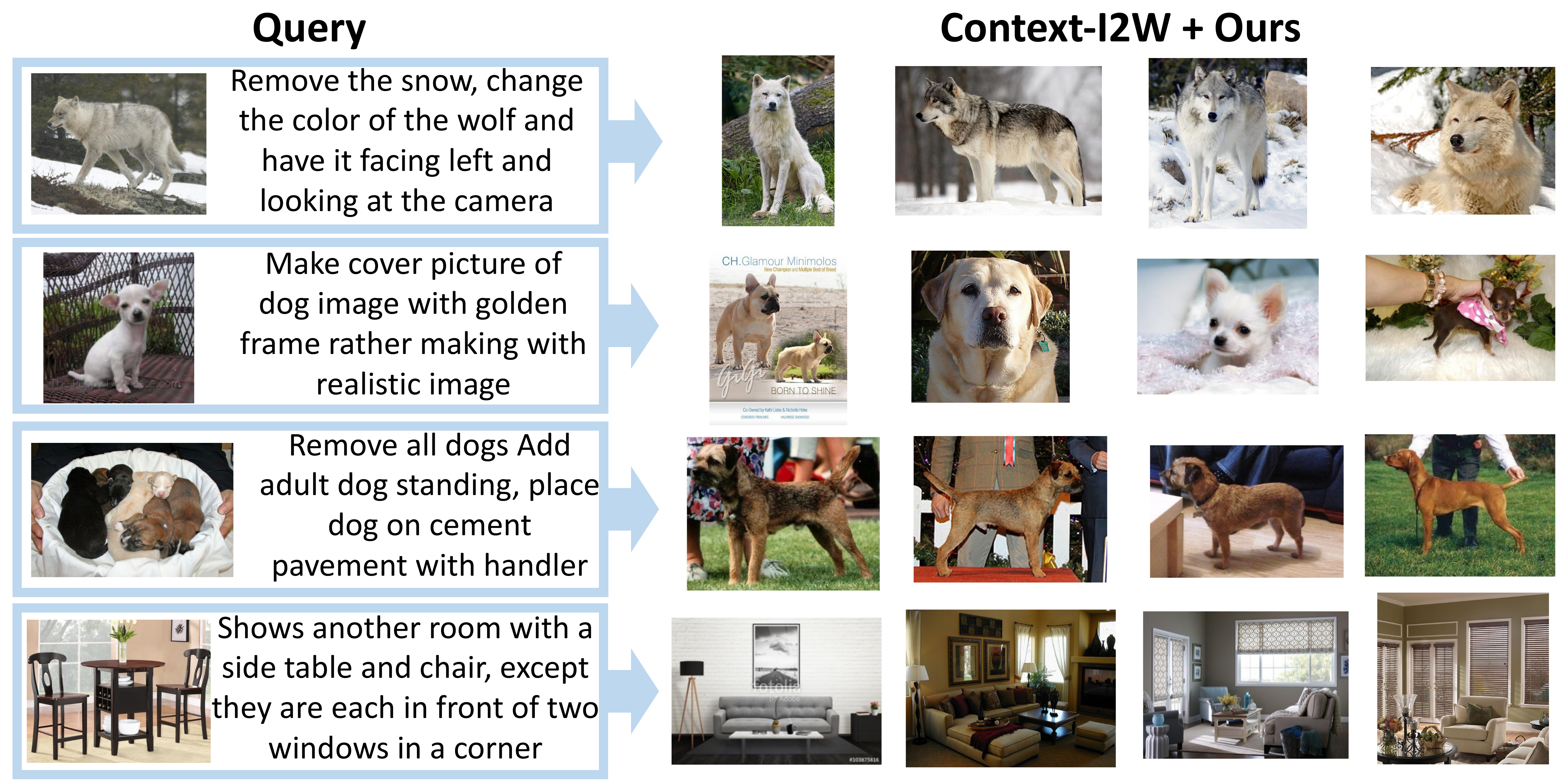}
    \caption{Visualization of common failure cases. Our method faces the challenge of handling complex and redundant manipulation descriptions. }
    \label{fig:fail}
\end{figure*}

\subsection{Analysis of the size of copped images.} 
In this section, we evaluated the effect of the width and height of the cropping box across four datasets, as shown in Figure \ref{fig:crop_ablation}. We found that the size of the cropped image plays a crucial role in performance, primarily due to how much of the context from the pseudo-reference images and pseudo-manipulation descriptions it captures. Specifically, a smaller cropping size range of \(16\sim32\) provides insufficient contextual overlap with target images, resulting in a decreased ability to learn the intended image-to-text mapping for refining visual details, leading to reduced performance. Conversely, larger cropping sizes of \(128\sim256\) include more context, but they complicate the learning process for pseudo-composed image retrieval, also diminishing performance. After testing various sizes, we determined that a cropping box size range of \(32\sim64\) strikes an optimal balance, enabling effective learning of the specific image-to-text mapping and pseudo-composed image retrieval, thus yielding the best results among the configurations tested.

\subsection{Effectiveness and Efficiency Analysis}
Our approach obtains significant improvement over four widely compared ZR-CIR tasks from 1.45\% to 4.17\% over SoTA models, without extra parameters and inference time. Due to additional image processing and pseudo triplets, our training time is $4$ hours longer than Pic2Word and Context-I2W and 30 minutes longer than SEARLE. It’s worth noting that our model using just 50\% of the pre-training data achieves comparable performance to SoTA models, as shown in our quantitative experiments.

\subsection{Visualization of Constructed Pseudo Triples}

In Figure \ref{fig:noise}, we present the pseudo-triplets generated using our 'Partial Cropped' and 'Style Transfer' strategies. The partial-cropped images include partial contexts relevant to their captions, enhancing the training for pseudo-composed image retrieval by supplementing the pseudo-manipulation descriptions. Although they differ stylistically, the style-transfer images maintain contextual similarities with the original images, a pattern that consistently appears. These visualizations demonstrate that our Pseudo Triplet Construction module effectively constructs reference images that enhance the contextual relevance of captions, facilitating attribute manipulation and domain conversion.

Furthermore, we showcase examples of noise samples and filtered pseudo-triplets, highlighting the efficiency of our 'Cropped Image Filtering' method in removing potential noise from pseudo triplets. This method not only aids in maintaining the quality of the training data but can also serve as a valuable preprocessing step during training, ensuring cleaner and more relevant data inputs.

\subsection{Discussion on Common Failure Cases}

Figure \ref{fig:fail} illustrates common failure scenarios for our Denoise-I2W model, particularly when dealing with complex and redundant manipulation descriptions. Notable challenges include simultaneous attribute and scene manipulation (row 1), object manipulation alongside domain conversion (row 2), and extensive manipulation involving multiple objects and scenes (row 3). These issues primarily stem from the limitations of the CLIP language encoder in comprehensively understanding and processing complex and redundant manipulation intentions for effective retrieval. Addressing these challenges by developing models that can better understand and handle complex and redundant manipulation intentions will be our future work.

\section{Conclusion}
In this paper, we propose a novel intention-based denoising image-to-word mapping method that unifies pre-training and inference goals, reduces redundant visual information, and focuses on the visual aspects of manipulation intention for accurate ZS-CIR. Denoise-I2W shows strong generalization ability and remarkably improves the best performance of existing approaches on four ZS-CIR tasks. It inspires the vision-to-language alignment mechanism by manipulating intent and impacting diverse vision and language applications. How to develop models for understanding complex and redundant manipulation intentions will be the future work.

\bibliographystyle{IEEEtran}
\bibliography{egbib}

\begin{thebibliography}{10}
\providecommand{\url}[1]{#1}
\csname url@samestyle\endcsname
\providecommand{\newblock}{\relax}
\providecommand{\bibinfo}[2]{#2}
\providecommand{\BIBentrySTDinterwordspacing}{\spaceskip=0pt\relax}
\providecommand{\BIBentryALTinterwordstretchfactor}{4}
\providecommand{\BIBentryALTinterwordspacing}{\spaceskip=\fontdimen2\font plus
\BIBentryALTinterwordstretchfactor\fontdimen3\font minus \fontdimen4\font\relax}
\providecommand{\BIBforeignlanguage}[2]{{%
\expandafter\ifx\csname l@#1\endcsname\relax
\typeout{** WARNING: IEEEtran.bst: No hyphenation pattern has been}%
\typeout{** loaded for the language `#1'. Using the pattern for}%
\typeout{** the default language instead.}%
\else
\language=\csname l@#1\endcsname
\fi
#2}}
\providecommand{\BIBdecl}{\relax}
\BIBdecl

\bibitem{datta2008image}
R.~Datta, D.~Joshi, J.~Li, and J.~Z. Wang, ``Image retrieval: Ideas, influences, and trends of the new age,'' \emph{ACM Computing Surveys}, vol.~40, no.~2, pp. 1--60, 2008.

\bibitem{vo2019composing}
N.~Vo, L.~Jiang, C.~Sun, K.~Murphy, L.-J. Li, L.~Fei-Fei, and J.~Hays, ``Composing text and image for image retrieval-an empirical odyssey,'' in \emph{Proceedings of the IEEE/CVF conference on computer vision and pattern recognition}, 2019, pp. 6439--6448.

\bibitem{Chen_2020_CVPR}
Y.~Chen, S.~Gong, and L.~Bazzani, ``Image search with text feedback by visiolinguistic attention learning,'' in \emph{Proceedings of the IEEE/CVF Conference on Computer Vision and Pattern Recognition}, 2020, pp. 3001--3011.

\bibitem{Saito_2023_CVPR}
K.~Saito, K.~Sohn, X.~Zhang, C.-L. Li, C.-Y. Lee, K.~Saenko, and T.~Pfister, ``Pic2word: Mapping pictures to words for zero-shot composed image retrieval,'' in \emph{Proceedings of the IEEE/CVF Conference on Computer Vision and Pattern Recognition}, 2023, pp. 19\,305--19\,314.

\bibitem{Liu_2021_ICCV}
Z.~Liu, C.~Rodriguez-Opazo, D.~Teney, and S.~Gould, ``Image retrieval on real-life images with pre-trained vision-and-language models,'' in \emph{Proceedings of the IEEE/CVF International Conference on Computer Vision}, October 2021, pp. 2125--2134.

\bibitem{Goenka_2022_CVPR}
S.~Goenka, Z.~Zheng, A.~Jaiswal, R.~Chada, Y.~Wu, V.~Hedau, and P.~Natarajan, ``Fashionvlp: Vision language transformer for fashion retrieval with feedback,'' in \emph{Proceedings of the IEEE/CVF Conference on Computer Vision and Pattern Recognition}, June 2022, pp. 14\,105--14\,115.

\bibitem{Baldrati_2022_CVPR}
A.~Baldrati, M.~Bertini, T.~Uricchio, and A.~Del~Bimbo, ``Effective conditioned and composed image retrieval combining clip-based features,'' in \emph{Proceedings of the IEEE/CVF Conference on Computer Vision and Pattern Recognition}, June 2022, pp. 21\,466--21\,474.

\bibitem{baldrati2023zero}
A.~Baldrati, L.~Agnolucci, M.~Bertini, and A.~Del~Bimbo, ``Zero-shot composed image retrieval with textual inversion,'' \emph{arXiv:2303.15247}, 2023.

\bibitem{tang2023contexti2w}
Y.~Tang, J.~Yu, K.~Gai, J.~Zhuang, G.~Xiong, Y.~Hu, and Q.~Wu, ``Context-i2w: Mapping images to context-dependent words for accurate zero-shot composed image retrieval,'' in \emph{Proceedings of the AAAI Conference on Artificial Intelligence}, vol.~38, no.~6, 2024, pp. 5180--5188.

\bibitem{radford2021learning}
A.~Radford, J.~W. Kim, C.~Hallacy, A.~Ramesh, G.~Goh, S.~Agarwal, G.~Sastry, A.~Askell, P.~Mishkin, J.~Clark, G.~Krueger, and I.~Sutskever, ``Learning transferable visual models from natural language supervision,'' in \emph{Proceedings of the International Conference on Machine Learning}, 2021, pp. 8748--8763.

\bibitem{Vo_2019_CVPR}
N.~Vo, L.~Jiang, C.~Sun, K.~Murphy, L.-J. Li, L.~Fei-Fei, and J.~Hays, ``Composing text and image for image retrieval - an empirical odyssey,'' in \emph{Proceedings of the IEEE/CVF Conference on Computer Vision and Pattern Recognition}, 2019, pp. 6439--6448.

\bibitem{gu2024lincir}
G.~Gu, S.~Chun, W.~Kim, , Y.~Kang, and S.~Yun, ``Language-only efficient training of zero-shot composed image retrieval,'' in \emph{Conference on Computer Vision and Pattern Recognition (CVPR)}, 2024.

\bibitem{zhang2024magiclens}
K.~Zhang, Y.~Luan, H.~Hu, K.~Lee, S.~Qiao, W.~Chen, Y.~Su, and M.-W. Chang, ``Magiclens: Self-supervised image retrieval with open-ended instructions,'' 2024.

\bibitem{agnolucci2024isearle}
L.~Agnolucci, A.~Baldrati, M.~Bertini, and A.~D. Bimbo, ``isearle: Improving textual inversion for zero-shot composed image retrieval,'' 2024.

\bibitem{xu2024set}
Y.~Xu, J.~Wei, Y.~Bin, Y.~Yang, Z.~Ma, and H.~T. Shen, ``Set of diverse queries with uncertainty regularization for composed image retrieval,'' \emph{IEEE Transactions on Circuits and Systems for Video Technology}, 2024.

\bibitem{mtcir}
J.~Chen and H.~Lai, ``Pretrain like you inference: Masked tuning improves zero-shot composed image retrieval,'' \emph{arXiv preprint arXiv:2311.07622}, 2023.

\bibitem{training-free-cir}
S.~Karthik, K.~Roth, M.~Mancini, and Z.~Akata, ``Vision-by-language for training-free compositional image retrieval,'' \emph{arXiv preprint arXiv:2310.09291}, 2023.

\bibitem{blip}
J.~Li, D.~Li, C.~Xiong, and S.~Hoi, ``Blip: Bootstrapping language-image pre-training for unified vision-language understanding and generation,'' in \emph{ICML}, 2022.

\bibitem{blip2}
J.~Li, D.~Li, S.~Savarese, and S.~Hoi, ``Blip-2: Bootstrapping language-image pre-training with frozen image encoders and large language models,'' in \emph{ICML}, 2023.

\bibitem{Zhou_2022_CVPR}
K.~Zhou, J.~Yang, C.~C. Loy, and Z.~Liu, ``Conditional prompt learning for vision-language models,'' in \emph{Proceedings of the IEEE/CVF Conference on Computer Vision and Pattern Recognition}, 2022, pp. 16\,816--16\,825.

\bibitem{song2022clip}
H.~Song, L.~Dong, W.-N. Zhang, T.~Liu, and F.~Wei, ``Clip models are few-shot learners: Empirical studies on vqa and visual entailment,'' 2022.

\bibitem{clip2fl}
J.~Shi, S.~Zheng, X.~Yin, Y.~Lu, Y.~Xie, and Y.~Qu, ``Clip-guided federated learning on heterogeneity and long-tailed data,'' in \emph{Proceedings of the AAAI Conference on Artificial Intelligence}, vol.~38, no.~13, 2024, pp. 14\,955--14\,963.

\bibitem{mmm}
J.~Shi, X.~Yin, Y.~Chen, Y.~Zhang, Z.~Zhang, Y.~Xie, and Y.~Qu, ``Multi-memory matching for unsupervised visible-infrared person re-identification,'' \emph{arXiv preprint arXiv:2401.06825}, 2024.

\bibitem{pmlr-v162-li22n}
J.~Li, D.~Li, C.~Xiong, and S.~Hoi, ``{BLIP}: Bootstrapping language-image pre-training for unified vision-language understanding and generation,'' in \emph{Proceedings of the 39th International Conference on Machine Learning}, 2022, pp. 12\,888--12\,900.

\bibitem{li2020oscar}
X.~Li, X.~Yin, C.~Li, P.~Zhang, X.~Hu, L.~Zhang, L.~Wang, H.~Hu, L.~Dong, F.~Wei \emph{et~al.}, ``Oscar: Object-semantics aligned pre-training for vision-language tasks,'' in \emph{European Conference on Computer Vision}, 2020, pp. 121--137.

\bibitem{zhang2021vinvl}
P.~Zhang, X.~Li, X.~Hu, J.~Yang, L.~Zhang, L.~Wang, Y.~Choi, and J.~Gao, ``Vinvl: Revisiting visual representations in vision-language models,'' in \emph{Proceedings of the IEEE/CVF Conference on Computer Vision and Pattern Recognition}, 2021, pp. 5579--5588.

\bibitem{10.1007/978-3-031-20044-1_32}
N.~Cohen, R.~Gal, E.~A. Meirom, G.~Chechik, and Y.~Atzmon, ````this is my unicorn, fluffy'': Personalizing frozen vision-language representations,'' in \emph{European conference on computer vision}, 2022, pp. 558--577.

\bibitem{Kumari_2023_CVPR}
N.~Kumari, B.~Zhang, R.~Zhang, E.~Shechtman, and J.-Y. Zhu, ``Multi-concept customization of text-to-image diffusion,'' in \emph{Proceedings of the IEEE/CVF Conference on Computer Vision and Pattern Recognition}, 2023, pp. 1931--1941.

\bibitem{mokady2021clipcap}
R.~Mokady, A.~Hertz, and A.~H. Bermano, ``Clipcap: Clip prefix for image captioning,'' 2021.

\bibitem{zhu2023visualize}
W.~Zhu, A.~Yan, Y.~Lu, W.~Xu, X.~E. Wang, M.~Eckstein, and W.~Y. Wang, ``Visualize before you write: Imagination-guided open-ended text generation,'' 2023.

\bibitem{tam2023simple}
\BIBentryALTinterwordspacing
D.~Tam, C.~Raffel, and M.~Bansal, ``Simple weakly-supervised image captioning via {CLIP}'s multimodal embeddings,'' in \emph{The AAAI-23 Workshop on Creative AI Across Modalities}, 2023. [Online]. Available: \url{https://openreview.net/forum?id=UMxeP-FuwyC}
\BIBentrySTDinterwordspacing

\bibitem{10.1007/978-3-319-10602-1_48}
T.-Y. Lin, M.~Maire, S.~Belongie, J.~Hays, P.~Perona, D.~Ramanan, P.~Doll{\'a}r, and C.~L. Zitnick, ``Microsoft coco: Common objects in context,'' in \emph{European Conference on Computer Vision}, D.~Fleet, T.~Pajdla, B.~Schiele, and T.~Tuytelaars, Eds., 2014, pp. 740--755.

\bibitem{deng2009imagenet}
J.~Deng, W.~Dong, R.~Socher, L.-J. Li, K.~Li, and L.~Fei-Fei, ``Imagenet: A large-scale hierarchical image database,'' in \emph{Computer Vision and Pattern Recognition}, 2009, pp. 248--255.

\bibitem{Hendrycks_2021_ICCV}
D.~Hendrycks, S.~Basart, N.~Mu, S.~Kadavath, F.~Wang, E.~Dorundo, R.~Desai, T.~Zhu, S.~Parajuli, M.~Guo, D.~Song, J.~Steinhardt, and J.~Gilmer, ``The many faces of robustness: A critical analysis of out-of-distribution generalization,'' in \emph{Proceedings of the IEEE/CVF International Conference on Computer Vision}, 2021, pp. 8340--8349.

\bibitem{Wu_2021_CVPR}
H.~Wu, Y.~Gao, X.~Guo, Z.~Al-Halah, S.~Rennie, K.~Grauman, and R.~Feris, ``Fashion iq: A new dataset towards retrieving images by natural language feedback,'' in \emph{Proceedings of the IEEE/CVF Conference on Computer Vision and Pattern Recognition}, 2021, pp. 11\,307--11\,317.

\bibitem{DBLP:conf/acl/SoricutDSG18}
P.~Sharma, N.~Ding, S.~Goodman, and R.~Soricut, ``Conceptual captions: A cleaned, hypernymed, image alt-text dataset for automatic image captioning,'' in \emph{Annual Meeting of the Association for Computational Linguistics}, 2018, pp. 2556--2565.

\bibitem{loshchilov2018decoupled}
I.~Loshchilov and F.~Hutter, ``Decoupled weight decay regularization,'' in \emph{International Conference on Learning Representations}, 2018.

\bibitem{brown2020language}
\BIBentryALTinterwordspacing
T.~Brown, B.~Mann, N.~Ryder, M.~Subbiah, J.~D. Kaplan, P.~Dhariwal, A.~Neelakantan, P.~Shyam, G.~Sastry, A.~Askell, S.~Agarwal, A.~Herbert-Voss, G.~Krueger, T.~Henighan, R.~Child, A.~Ramesh, D.~Ziegler, J.~Wu, C.~Winter, C.~Hesse, M.~Chen, E.~Sigler, M.~Litwin, S.~Gray, B.~Chess, J.~Clark, C.~Berner, S.~McCandlish, A.~Radford, I.~Sutskever, and D.~Amodei, ``Language models are few-shot learners,'' in \emph{Advances in Neural Information Processing Systems}, H.~Larochelle, M.~Ranzato, R.~Hadsell, M.~Balcan, and H.~Lin, Eds., vol.~33.\hskip 1em plus 0.5em minus 0.4em\relax Curran Associates, Inc., 2020, pp. 1877--1901. [Online]. Available: \url{https://proceedings.neurips.cc/paper_files/paper/2020/file/1457c0d6bfcb4967418bfb8ac142f64a-Paper.pdf}
\BIBentrySTDinterwordspacing

\bibitem{delmas2022artemis}
G.~Delmas, R.~S. Rezende, G.~Csurka, and D.~Larlus, ``{ARTEMIS}: Attention-based retrieval with text-explicit matching and implicit similarity,'' in \emph{International Conference on Learning Representations}, 2022.

\end{thebibliography}
\vfill
\vspace{-30pt}
\begin{IEEEbiography}[{\includegraphics[width=1in,height=1.25in,clip,keepaspectratio]{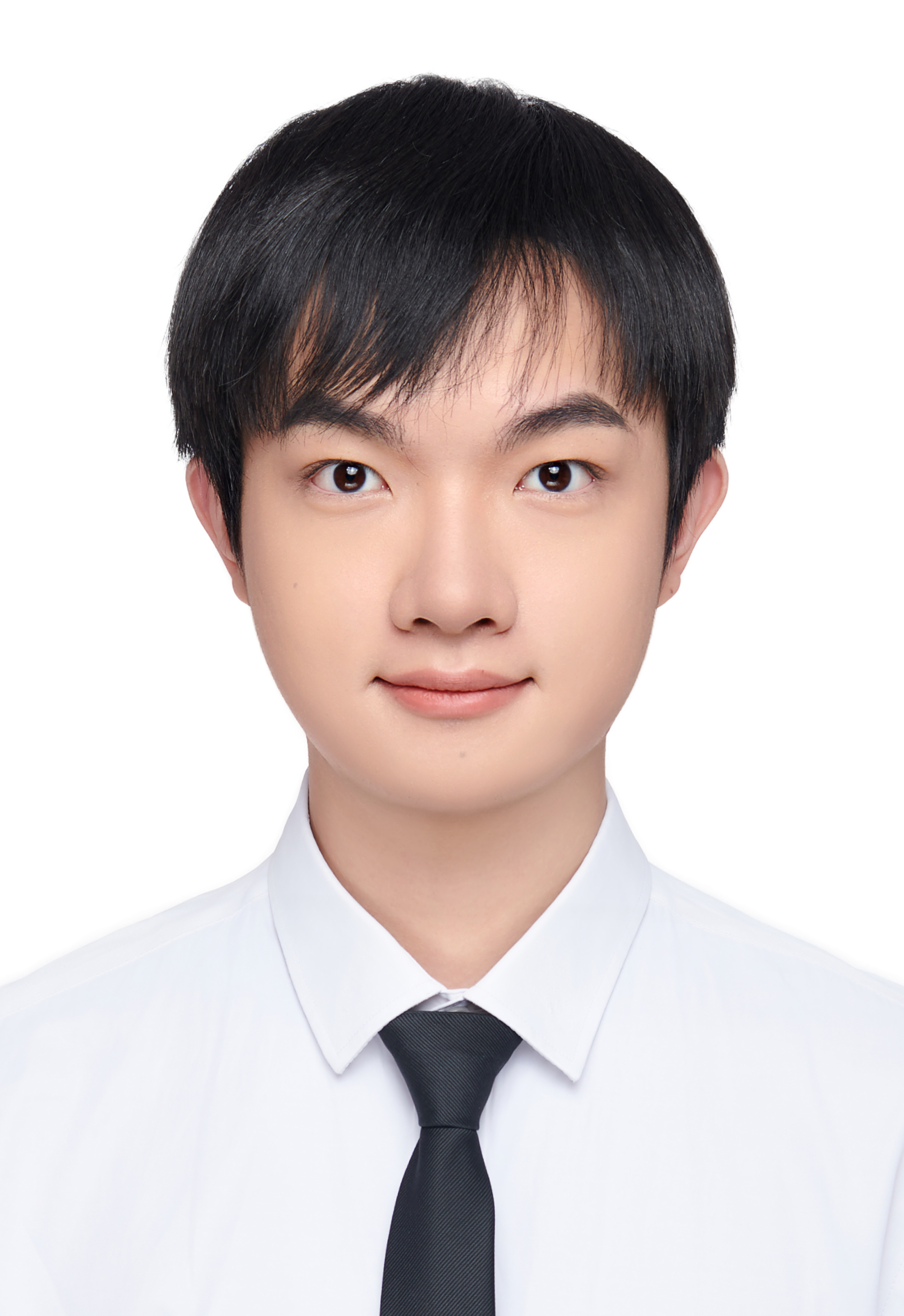}}]{Yuanmin Tang}
is currently studying for a Ph.D. degree in the Institute of Information Engineering, Chinese Academy of Sciences, Beijing, China. Yuanmin Tang received his B.S. degree in Computer Science from Henan University, China, in 2017. His research interests mainly focus on cross-model alignment, including composed image retrieval, cross-model sponsored search, etc. His work has been published in prestigious journals and conferences such as AAAI.
\end{IEEEbiography}

\begin{IEEEbiography}[{\includegraphics[width=1in,height=1.25in,clip,keepaspectratio]{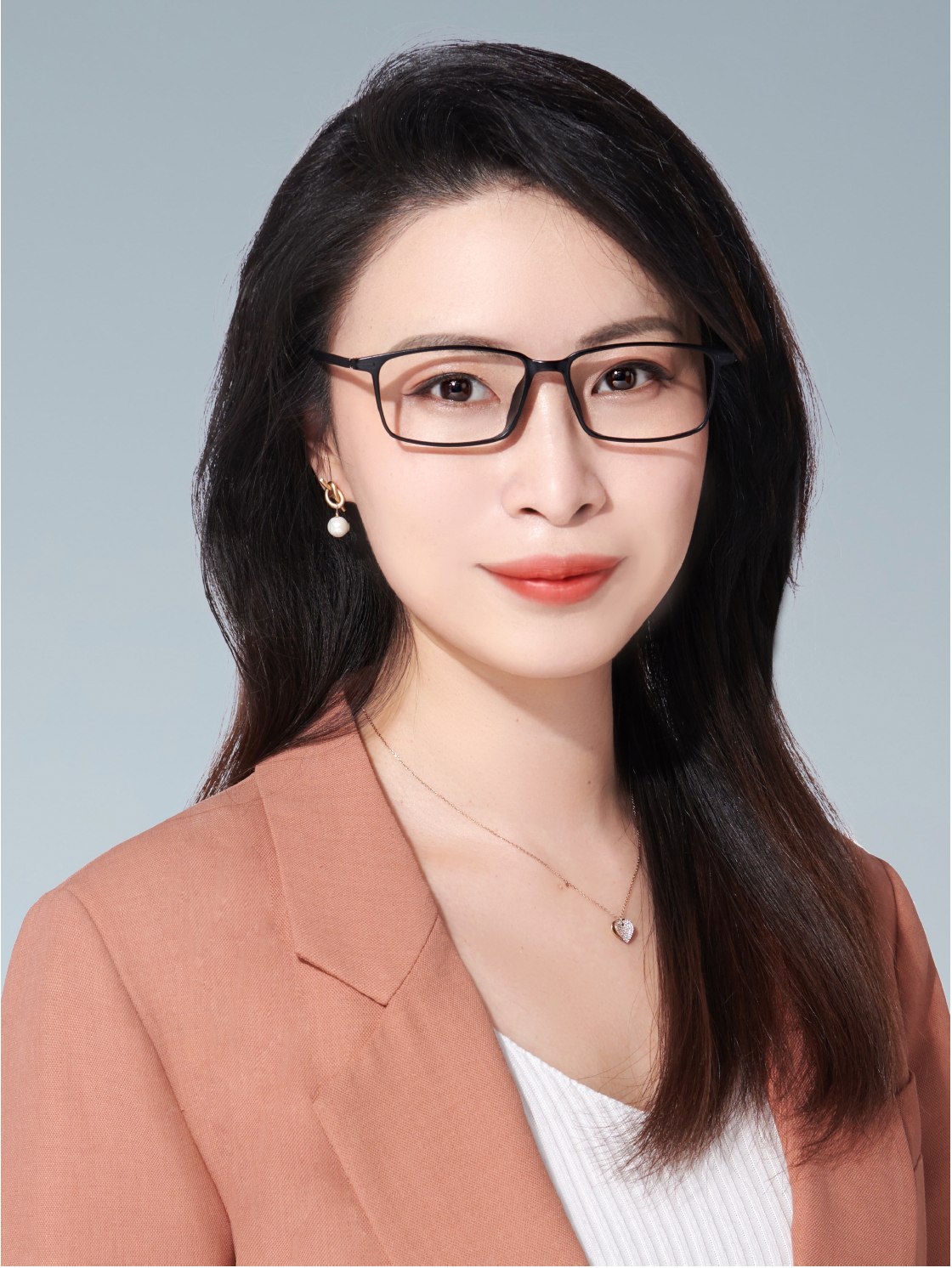}}]{Jing Yu}
is an Associate Professor at the School of Information Engineering, Minzu University of China. IEEE Computer Society Smart Computing Special Technical Community awarded her a Early-Career Award in 2023. Jing Yu received her B.S. degree in Automation Science from Minzu University, China, in 2011, and got her M.S. degree in Pattern Recognition from Beihang University, China in 2014. She recieved her Ph.D. degree in the University of Chinese Academy of Sciences, China, in 2019. Currently, she is working on the vision-language problem. She mainly focuses on cross-modal understanding, including visual question answering, cross-modal information retrieval, scene graph generation, etc. She has published more than 70 papers in prestigious conferences and journals, such as CVPR, AAAI, IJCAI. 
\end{IEEEbiography}

\begin{IEEEbiography}[{\includegraphics[width=1in,height=1.25in,clip,keepaspectratio]{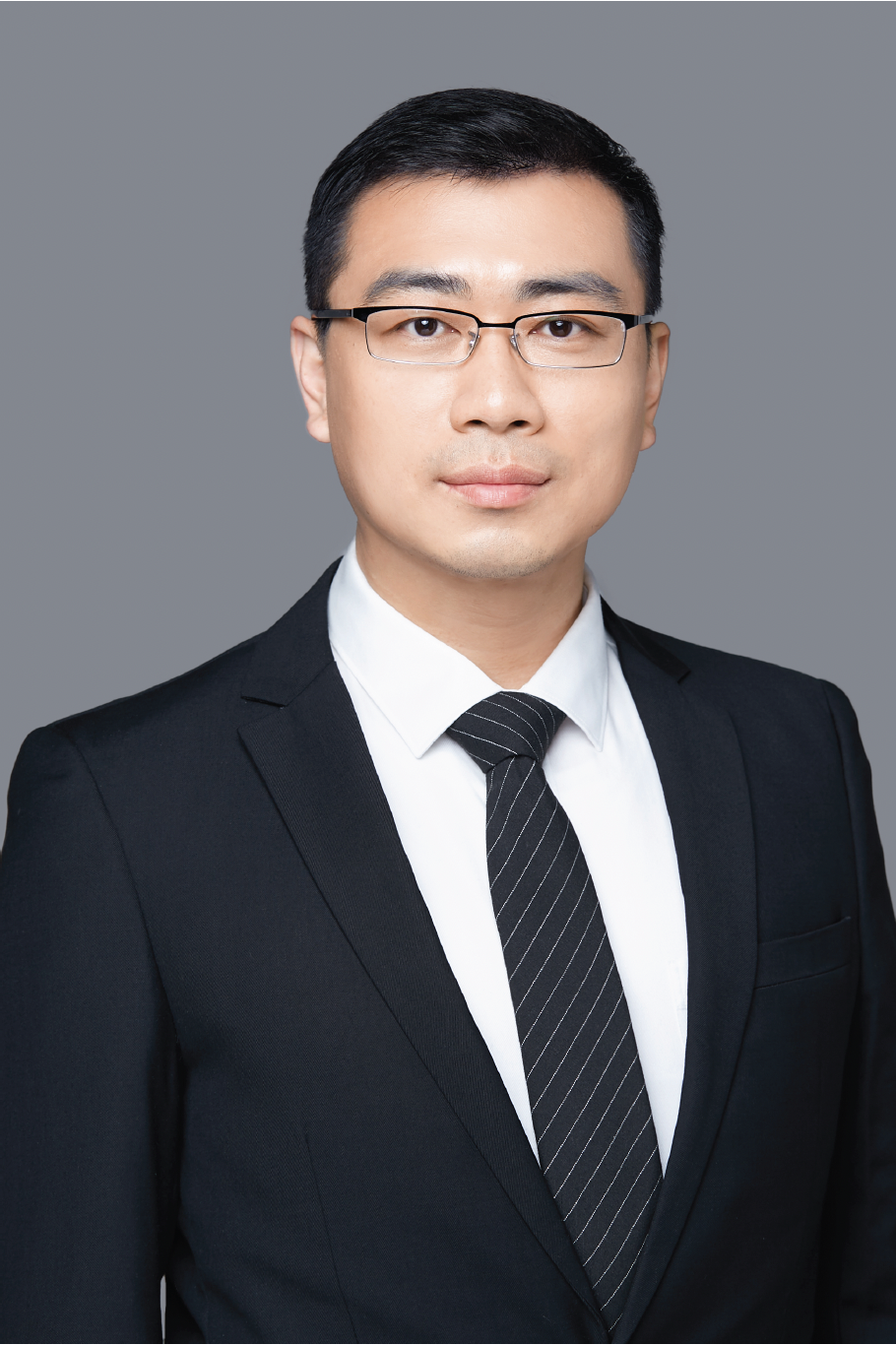}}]{Keke Gai}
received the B.Eng. degree majored in automation, from Nanjing University of Science and Technology, Nanjing, China, in 2004, the M.E.T. (Master’s of Educational Technology) degree in educational technology from the University of British Columbia, Vancouver, BC, Canada, in 2010, and the Ph.D. degree in computer science from Pace University, New York, NY, USA. He is currently a Professor at the School of Cyberspace Science and Technology, Beijing Institute of Technology, Beijing, China. He has published 4 books and over 160 peer-reviewed journal/conference papers, including 10 best paper awards. He serves as an Editor-in-Chief of the journal Blockchains and Associate Editors for several decent journals, including IEEE Transactions on Dependable and Secure Computing, Journal of Parallel and Distributed Computing, etc. He also serves as a co-chair of IEEE Technology and Engineering Management Society’s Technical Committee on Blockchain and Distributed Ledger Technologies, a Standing Committee Member at China Computer Federation - Blockchain Committee, a Secretary-General at IEEE Special Technical Community in Smart Computing. His research interests include cyber security, blockchain, privacy-preserving computation, decentralized identity, and artiﬁcial intelligence security.
\end{IEEEbiography}

\begin{IEEEbiography}[{\includegraphics[width=1in,height=1.25in, clip,keepaspectratio]{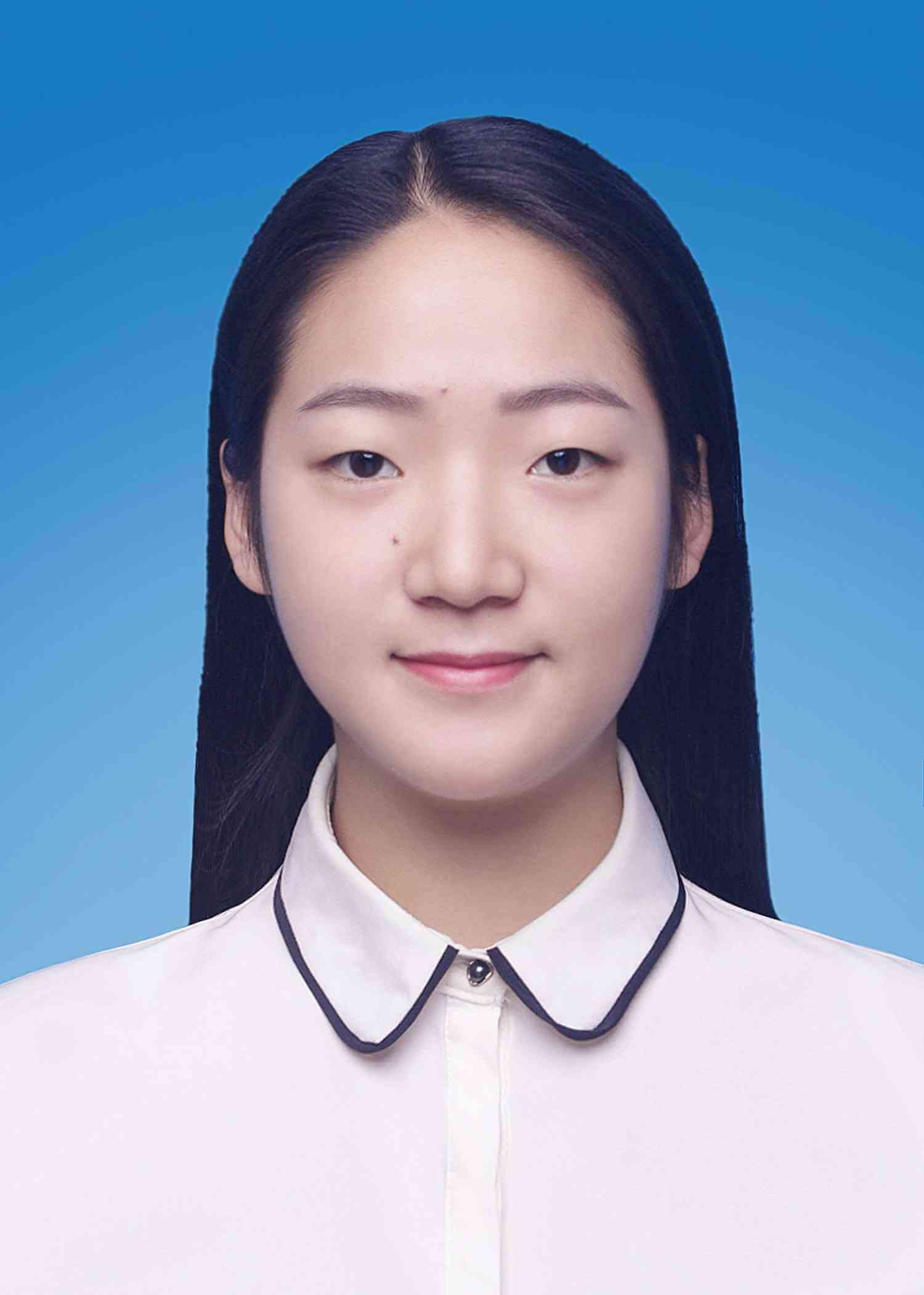}}]{Jiamin Zhuang} is currently studying for a Ph.D. degree at the Institute of Information Engineering, Chinese Academy of Sciences, Beijing, China. Jiamin Zhuang received her B.S. degree in network engineering from Henan University, China, in 2016. Her research interests mainly focus on cross-modal retrieval.
\end{IEEEbiography}

\begin{IEEEbiography}[{\includegraphics[width=1in,height=1.25in,clip,keepaspectratio]{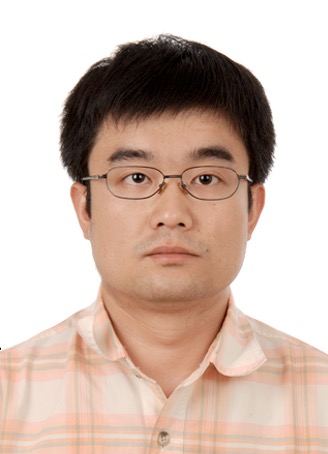}}]{Gaopeng Gou}
received the B.E., M.Eng., and Ph.D. degrees from Beihang University, China, in 2005, 2008, and 2014, respectively. He is currently a Senior Engineer and Full Professor at the Institute of Information Engineering, Chinese Academy of Sciences, Beijing, China. His research interests include network security and network anomaly detection.
\end{IEEEbiography}

\begin{IEEEbiography}[{\includegraphics[width=1in,height=1.25in,clip,keepaspectratio]{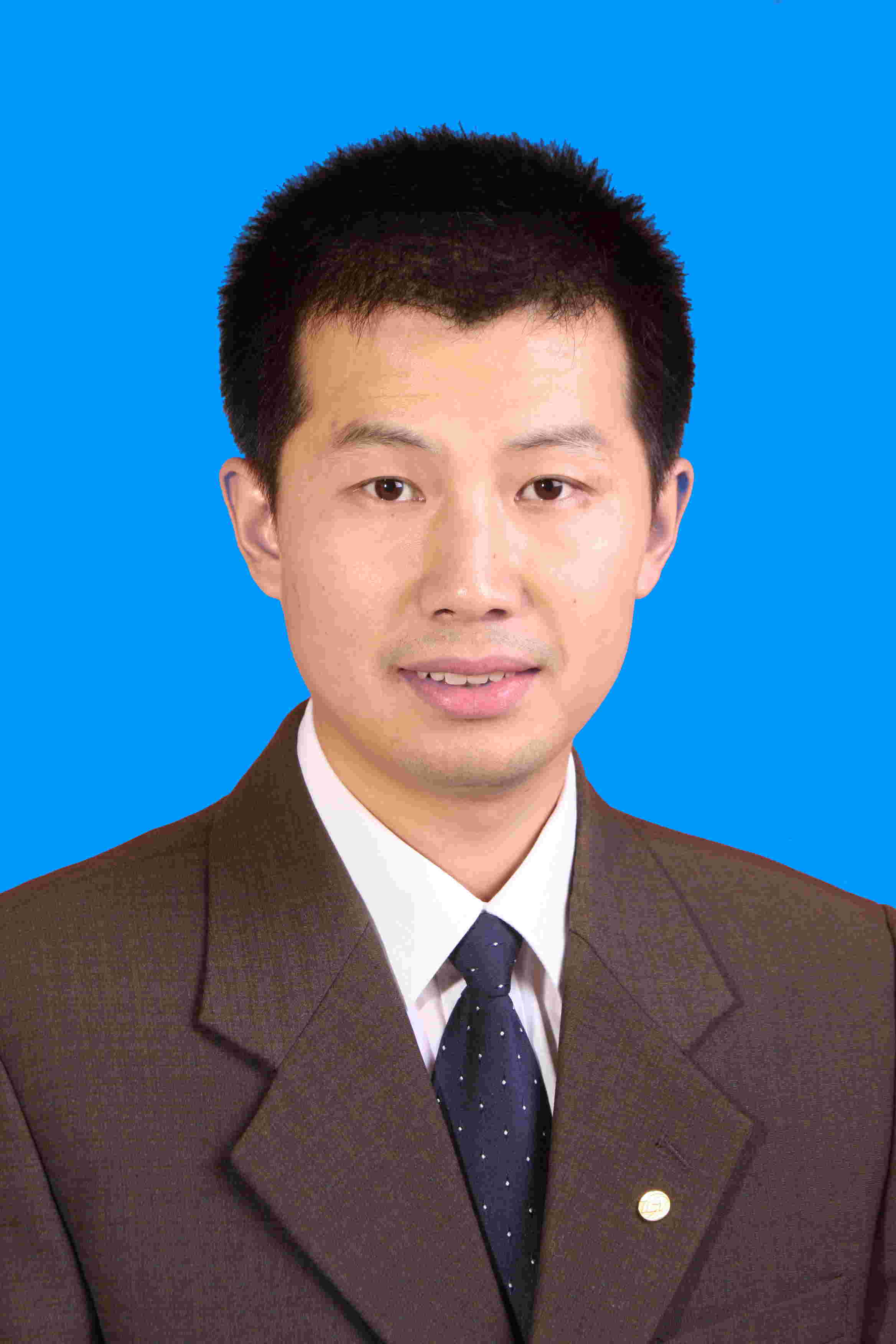}}]{Gang Xiong}
is currently a Professor at the Institute of Information Engineering, Chinese Academy of Sciences, China. He has authored more than 80 papers in refereed journals and conference proceedings. His research interests include network and information security. He is a member of the 3rd Communication Security Technical Committee of China Institute of Communications.
\end{IEEEbiography}

\begin{IEEEbiography}[{\includegraphics[width=1in,height=1.25in,clip,keepaspectratio]{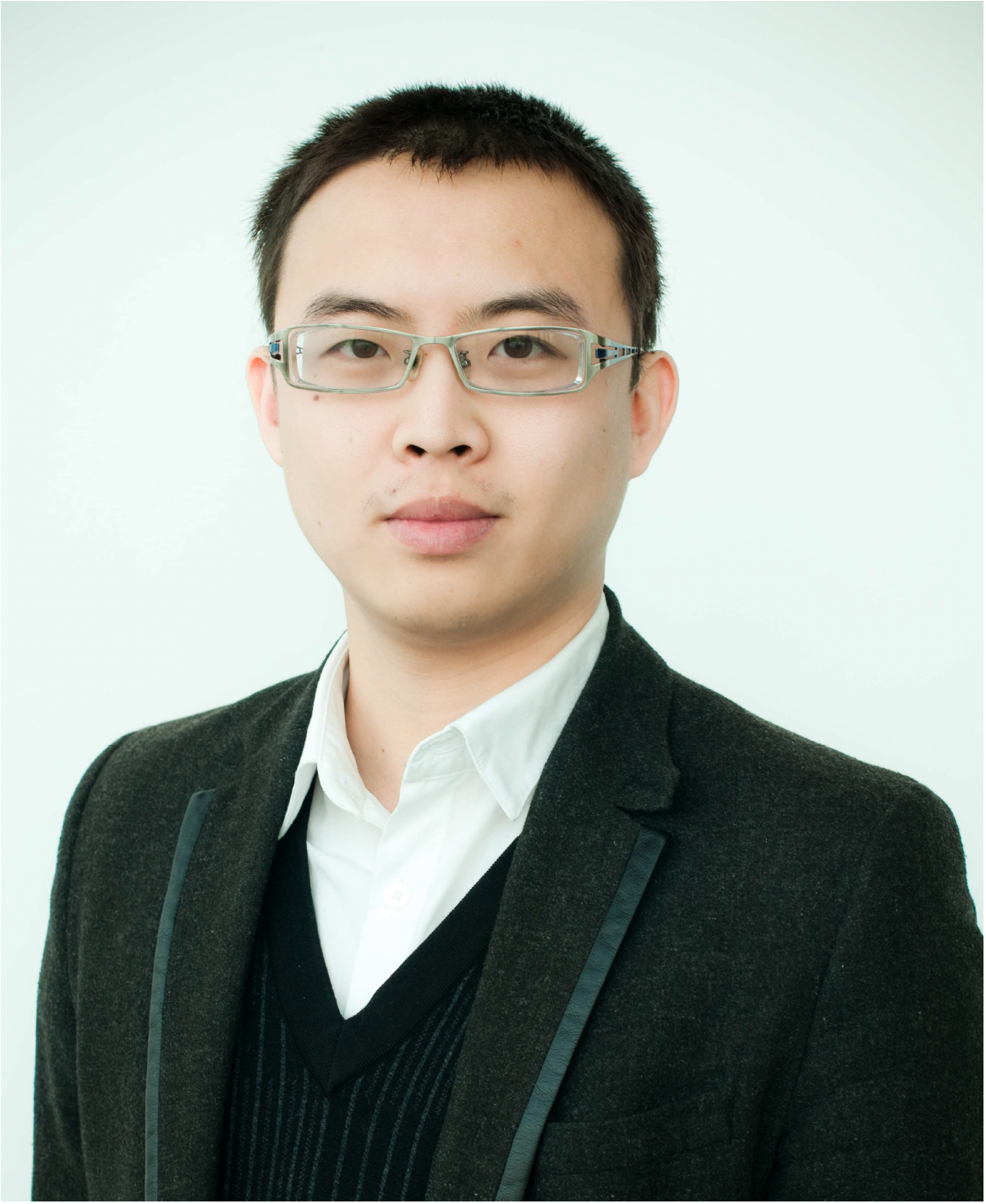}}]{Qi Wu}
 is an Associate Professor at the University of Adelaide and was the ARC Discovery Early Career Researcher Award (DECRA) Fellow between 2019-2021. He is the Director of Vision-and-Language at the Australia Institute of Machine Learning. Australian Academy of Science awarded him a J G Russell Award in 2019. He obtained his PhD degree in 2015 and MSc degree in 2011, in Computer Science from the University of Bath, United Kingdom. His research interests are mainly in computer vision and machine learning. Currently, he is working on the vision-language problem, and he is primarily an expert in image captioning and visual question answering (VQA). He has published more than 100 papers in prestigious conferences and journals, such as TPAMI, CVPR, ICCV, ECCV. He is also the Area Chair for CVPR and ICCV.
\end{IEEEbiography}

\vspace{11pt}


\end{document}